\documentclass[runningheads]{llncs}
\usepackage{color}
\usepackage{microtype}
\usepackage{graphicx}
\usepackage{subfigure}
\usepackage{booktabs} %
\usepackage{comment}
\usepackage{multirow}
\usepackage{textcomp}
\usepackage{siunitx}
\usepackage{float}
\usepackage{enumitem}
\usepackage{adjustbox}
\usepackage{capt-of}
\newsavebox{\tempbox}

\usepackage{hyperref}

\usepackage{algorithm}%
\usepackage{algorithmic}

\usepackage{array}
\newcolumntype{?}{!{\vrule width 1pt}}

\usepackage{xcolor}

\usepackage{amsmath,amsfonts}
\usepackage{amssymb,amsopn}
\usepackage{bm} %
\usepackage{dsfont}
\DeclareRobustCommand\onedot{\futurelet\@let@token\@onedot}
\def\onedot{. } %

\makeatletter
        
        \def\env@closedcases{%
            \let\@ifnextchar\new@ifnextchar
            \left\{
            \def\arraystretch{1.2}%
            \array{@{}l@{\quad}l@{}}%
        }
\makeatother

\def\eg{\textit{e.g}\onedot} 
\def\ie{\textit{i.e}\onedot}

\def\etal{\textit{et al}\onedot}

\DeclareMathOperator*{\argmax}{arg\,max}

\DeclareMathOperator*{\argmin}{arg\,min} %

\usepackage[algo2e]{algorithm2e}

\SetKwIF{If}{ElseIf}{Else}{if~(\endgraf}{\endgraf)~then}{else if}{else}{end if}%

\begin{document}
\pagestyle{headings}
\mainmatter
\def\ECCVSubNumber{1912}  %

\title{Imbalanced Continual Learning with Partitioning Reservoir Sampling}

\titlerunning{ECCV-20 submission ID \ECCVSubNumber}
\authorrunning{ECCV-20 submission ID \ECCVSubNumber}
\author{Anonymous ECCV submission}
\institute{Paper ID \ECCVSubNumber}
\titlerunning{Imbalanced Continual Learning}
\author{Chris Dongjoo Kim \and
Jinseo Jeong \and
Gunhee Kim}

\authorrunning{C.D. Kim et al.}
\institute{Neural Processing Research Center, Seoul National University, Seoul, Korea
\email{\{cdjkim,jinseo\}@vision.snu.ac.kr, gunhee@snu.ac.kr}\\
\url{http://vision.snu.ac.kr/projects/PRS}}
\maketitle

\begin{abstract}
    Continual learning from a sequential stream of data is a crucial challenge for machine learning research. 
    Most studies have been conducted on this topic under the single-label classification setting along with an assumption of balanced label distribution.
This work expands this research horizon towards multi-label classification. In doing so, we identify unanticipated adversity innately existent in many multi-label datasets, 
the \textit{long-tailed} distribution. %
We jointly address the two independently solved problems, Catastropic Forgetting and the long-tailed label distribution
by first empirically showing a new challenge of destructive forgetting of the minority concepts on the tail.
    Then, we curate two benchmark datasets, \textit{COCOseq} and \textit{NUS-WIDEseq}, that allow the study of both \textit{intra}- and \textit{inter}-task imbalances. 
    Lastly, we propose a new sampling strategy for replay-based approach named \textit{Partitioning Reservoir Sampling} (PRS), which allows the model to maintain a balanced knowledge of both head and tail classes.
    We publicly release the dataset and the code in our project page.
\end{abstract}

\section{Introduction}
\label{sec:intro}

Sequential data streams are among the most natural forms of input for intelligent agents abiding the law of time.
Recently, there has been much effort to better learn from these types of inputs, termed \textit{continual learning} in machine learning research. %
Specifically, there have been many ventures into but not limited to single-label text classification~\cite{autume19neurips}, question answering~\cite{autume19neurips}, language instruction and translation~\cite{li20iclr},
object detection~\cite{shmelkov17iccv,liu19arxiv}, captioning~\cite{nguyen19arxiv} and even video representation learning~\cite{parisi17,parisi18}.
Surprisingly, we have yet to see continual learning for  multi-label classification, %
a more general and practical form of classification tasks since most real-world data are typically associated with several semantic concepts.

In order to study continual learning for multi-label classification, the first job would be to construct a research benchmark for it.
We select two of the most popular multi-label datasets, MSCOCO~\cite{lin14eccv} and NUS-WIDE~\cite{chua09acm}, and tailor them into
a sequence of mutually exclusive tasks, \textit{COCOseq} and \textit{NUS-WIDEseq}.
In the process, we recognize that large-scale multi-label datasets inevitably follow a \textit{long-tailed} distribution
where  a small number of categories contain a large number of samples while most have only a small amount of samples.
This naturally occurring phenomenon is widely observed in vision and language datasets~\cite{reed01,newman05,van17devil},
with a whole other branch of machine learning that has focused solely on this topic.
Consequently, to effectively perform continual learning on multi-label data, two major obstacles should be overcome simultaneously:
(i) the infamous \textit{catastrophic forgetting} problem~\cite{mccloskey89,ratcliff90,french99} and
(ii) the long-tailed distribution problem~\cite{chen04using,he08learning,wang17nips,liu19large}, which we jointly address in this work.

We adopt the replay-based approach~\cite{lopez17,hayes19memory,aljundi19gradient,shin17,rolnick19,lesort19b} to tackle continual learning, which explicitly stores the past experiences into a memory or a generative model, and rehearses them back with the new input samples. 
Although there also exists the prior-focused (\ie regularization-based)~\cite{kirkpatrick17ewc,zenke17,aljundi19selfless} and expansion-based methods~\cite{rusu16,yoon18},
the replay-based approaches have often shown superior results in terms of performance and memory efficiency.
Specifically, the replay memory with \textit{reservoir sampling}~\cite{vitter85} has been a
strong baseline, especially in the task-free continual setting~\cite{aljundi19gradient,lee20iclr}
that does not require explicit task labels during the training nor test phase. It is an optimistic avenue of continual learning that we also undertake.

To conclude the introduction, we outline the contributions of this work:
\begin{enumerate}[label=\Roman*.]
\item To the best of our knowledge, this is the first work to tackle the continual learning for multi-label classification.
To this end, we reveal that it is critical to correctly address the intra- and inter-task imbalances along with the prevailing catastrophic forgetting problem of continual learning.
\item For the study of this new problem, we extend the existing multi-label datasets into their continual versions called \textit{COCOseq} and \textit{NUS-WIDEseq}.
\item We propose a new replay method named \textit{Partitioning Reservoir Sampling} (PRS) for continual learning in heterogeneous and long-tailed data streams. %
We discover that the key to success is to allocate a sufficient portion of memory to the moderate and minority classes to retain a balanced knowledge of  present and past experiences. %
\end{enumerate}

\section{Motivation: Fatal Forgetting on the Tail Classes}
\label{sec:motivation}

The long-tailed data distribution is both an enduring and pervasive problem in machine learning~\cite{japkowicz02ida,he08learning}, 
as most real-world datasets are inherently imbalanced~\cite{reed01,newman05,zhu14cvpr,van17devil,ouyang16cvpr}.
Wang \etal \cite{wang17nips}, for example, stated that minimizing the skew in the data distribution by collecting more examples in the tail classes is an arduous task and
even if one manages to balance them along one dimension, they can become imbalanced in another.

\begin{figure}[t!]
    \begin{center}
    \includegraphics[width=\columnwidth]{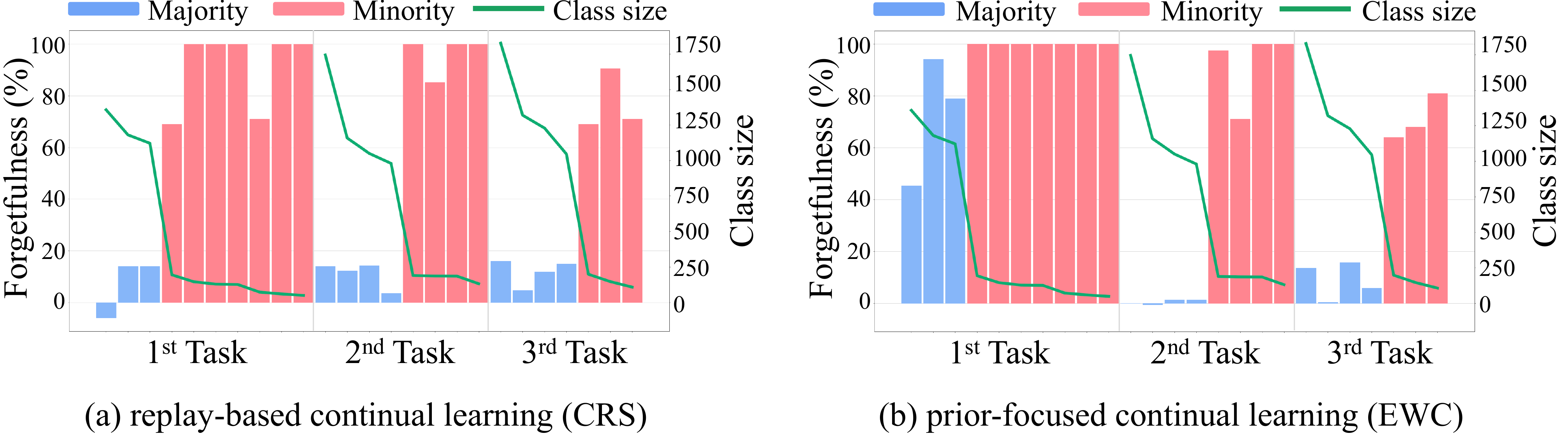}
    \caption{The forgetfulness of the majority and minority classes for two popular continual learning approaches over three sequential tasks.
     We test (a) replay-based reservoir sampling~\cite{brahma18subset,rolnick19,lopez17} and (b) EWC~\cite{kirkpatrick17ewc} with a shared output head and a memory size of 1000.
     We measure the forgetfulness using the metric proposed in \cite{chaudhry18eccv} (Higher is more forgetful). 
     The green line indicates the size of each class.
     More severe forgetting occurs for the minority classes in each task.}
    \label{fig:imbalance_problem}
    \end{center}
\end{figure}

We point out that the long-tailed distribution further aggravates the problem in continual learning, as a destructive amount of forgetting occurs on the tail classes.
We illustrate this with experiments on two existing continual learning approaches:
a prior-focused EWC~\cite{kirkpatrick17ewc} and a replay-based reservoir sampling~\cite{brahma18subset,rolnick19,lopez17}.
The experiments are carried out in an online setting on our \textit{COCOseq} dataset, whose detail will be presented in section \ref{sec:dataset}. %
Figure~\ref{fig:imbalance_problem} shows the results.
We plot the forgetting metric proposed in \cite{chaudhry18eccv}, which measures the difference between the peak performance and the performance at the end of the sequence.
For illustrative purposes, we sort the classes per task in decreasing order of the number of classes.
In both approaches, the minority (tail) classes experience more forgetting compared to the majority (head) classes.
We observe that the imbalance of sample distribution in the memory causes this phenomenon in accordance with the input distribution,
as we will further discuss in Figure~\ref{fig:buffer_distribution}.

\section{Approach}
\label{sec:approach}

The goal of this work is to overcome two inevitable obstacles of multi-label task-free continual learning: (i) catastrophic forgetting and (ii) long-tail input data distribution.
Since we adopt the replay-based approach, we focus on a new sampling strategy to reserve past experiences into a fixed memory.
We first clarify the problem (section \ref{sec:problem}), and discuss conventional reservoir sampling (section \ref{sec:crs}) and their fundamental limitations in this context (\ref{sec:prob_imbal}).
Finally, we propose our sampling method named \textit{Partitioning Reservoir Sampling} (section \ref{sec:prs}).

\subsection{Problem Formulation}
\label{sec:problem}

We formulate our multi-label task-free continual learning as follows.
The input is a data stream $S$, which consists of an unknown set of data points $(x, y)$, where $y$ is a multi-hot label vector representing $k$ arbitrary number of classes.
Except for the datapoint $(x, y)$ that enters in an online-manner, no other information (\eg the task boundaries or the number of classes) is available even during training.
Given an input stream, the goal of the model is to allocate the fixed memory $\mathcal{M}$ with a size of $m$: $\sum_{i=1}^{u} m_i \leq m$,
where $m_i$ denotes the partitioned memory size for class $c_i$, and $u$ is the unique number of classes observed so far at time $t$.

\subsection{Conventional Reservoir Sampling}
\label{sec:crs}

Conventional reservoir sampling (CRS)~\cite{vitter85} maintains a fixed memory that uniformly samples from an input data stream.
It is achieved by assigning a sampling probability of $m/n$ to each datapoint where $m$ is the memory size and $n$ is the total number of samples seen so far.
CRS is used as a standard sampling approach for task-free continual learning~\cite{rolnick19,chaudhry19iclr,riemer19iclr,lopez17},
since it does not require any prior information of the inputs but still attains an impressive performance~\cite{chaudhry19arxiv}.
However, its strength to uniformly represent the input distribution becomes its Achilles-heel in a long-tailed setting
as the memory distribution also becomes long-tailed, leading to the realm of problems experienced in imbalanced training.

\subsection{Fundamental Problems in Imbalanced Learning}
\label{sec:prob_imbal}

Imbalanced data induce severe issues in learning that are primarily attributed to gradient dominance and under-representation of the minority~\cite{krawczyk16,dong19tpami,zhu14cvpr,van17devil}.

(1) \textbf{Gradient dominance.}
The imbalance in the minibatch causes the majority classes dominating the gradient updates, which ultimately lead to the neglect of the minority classes.

(2) \textbf{Under-representation of the minority.}
Mainly due to the lack of data, the minority classes are much under-represented within the learned features relative to the majority~\cite{yin19cvpr,dong19tpami}.
We empirically confirm this in Figure~\ref{fig:t_sne}, where the minority classes do not formulate a discernable pattern in the feature space but are sparsely distributed by conventional methods.

There have been data processing or algorithmic approaches to tackle these problems by promoting balance during training. 
Data processing methods such as oversampling or undersampling~\cite{batista04,buda18,ouyang16cvpr} explicitly simulate the input balance,
while cost-sensitive approaches~\cite{huang16learning,wang17nips,cui19cvpr} adjust the update via regularizing the objective.
More aggressively, there have also been directions that populate the minority samples via generation to avoid overfitting in the minority~\cite{chawla02smote,maciejewski11local,douzas18effective}.
Most research in imbalanced learning shares the consensus that the \textit{balance} during training is critical to success,
which is the underlying emphasis on the design of our algorithm.

\subsection{Partitioning Reservoir Sampling}
\label{sec:prs}
\begin{figure}[t!]
    \begin{center}
    \includegraphics[width=\textwidth]{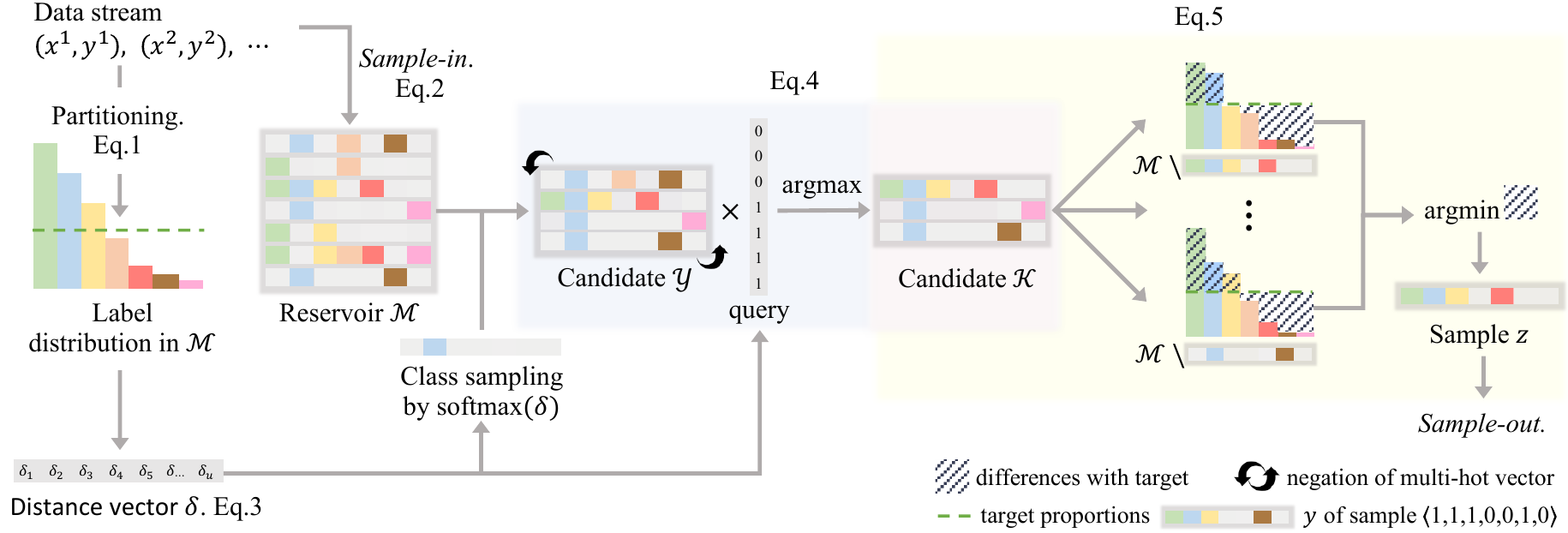}
    \caption{
    \textbf{Overview of Partitioning Reservoir Sampling.} Based on the current data stream statistics,
    the target partition ratios are first obtained based on Eq.~\ref{eq:partition}.
    We maintain the memory by iterating between the processes of \textit{sample-in} and {sample-out}.
    The sample-in decides whether a new datapoint is stored into the memory or not, while the sample-out selects which example is removed from the memory.
    If the model allows to sample-in the datapoint, 
    the algorithm traverses a process of candidate selection to sample-out an example by selecting the one that advances the memory towards the target partitions the most.
    }
    \label{fig:algorithm}
    \end{center}
\end{figure}

Since a continual replay algorithm has no information about future input, the memory must maintain  well-rounded knowledge in an online manner.
To that end, we provide an online memory maintenance algorithm called \textit{Partitioning Reservoir Sampling} (PRS) 
that consists of two fundamental operations: \textit{partition} and \textit{maintenance}.
The PRS is overviewed in Figure~\ref{fig:algorithm} and Algorithm~\ref{alg:prs}.

\textbf{The Partition.} During the training phase, the model only has access to the stream of data $(x, y)$.
While it is impractical to store all examples, caching the running statistics is a more sensible alternative.
Thus, the model uses the running class frequency to set the target proportion of classes in the memory.
This is achieved by a variant of the proportional allocation~\cite{carroll70power,fellegi88power,bankier88power}:  
\begin{align}
\label{eq:partition}
p_i = \frac{n_i^\rho}{\sum_j n_j^\rho}, %
\end{align}
where $\rho$ is a power of allocation, and $n_i$ is the running frequency of class $i$.
At $\rho=0$, all classes are equally allocated, which may be the most favorable scenario for the minority as it shares the same amount of memory with the majority.
At $\rho=1$, classes are allocated proportionally to their frequencies, which is identical to the conventional sampling in section \ref{sec:crs}.
$\rho$ is chosen a value between 0 and 1 to compromise between the two extremes.
For a given $\rho$, we can define the \textit{target partition quota} for class $i$ as $m_i=m \cdot p_i$ where $p_i$ is defined by Eq.~\ref{eq:partition}.
Collectively, the target partition is represented as a vector $\mathbf m =[m_1, \cdots, m_u]$, whose sum is $m$.
We will explore the effect of $\rho$ in Figure~\ref{fig:q_dist}.

\textbf{The Maintenance.}
The goal of maintenance is to allow every \textit{class} $i$ (not every sample as in CRS) to have a fair
chance of entering the memory according to the target partition $m_i $.
To maintain a well-rounded knowledge of the past and present experiences,
we iterate between the processes of \textit{sample-in} and \textit{sample-out}.
The sample-in decides whether a new input datapoint is reserved into the memory or not, while
the sample-out selects which example is removed from the memory when it becomes full and new samples continue to enter.

1) \textit{Sample-in}. %
For an incoming datapoint, we assign a sampling probability $s$ to be reserved in the memory.
We compute $s$ with two desiderata: (i)  it needs to comply with the target partition, and (ii) for better balancing, it is biased towards the minority classes with respect to the current running statistics.
\begin{align}
s = \sum_{i \in \{i,...,u\}} \frac{m_i}{n_i} \cdot w_i,\quad  \text{where } w_i = \frac{y_i e^{-n_i}}{\sum_{j=1} y_j e^{-n_j}} \label{eq:sample-in}
\end{align}
where $u$ is the unique number of classes observed, $n_i$ is the running frequency of class $i$, and $y_i$ is the datapoint's multi-hot vector value for class $i$.
$w_i$ is the normalized weight computed by the softmax of the negative running frequency of the classes.
This formulation allows to bias $w_i$ strongly towards the minority. %

2) \textit{Sample-out}.
When the memory $\mathcal M$ is full and new samples continue to enter, we need to sample out an example from the memory while striving towards the target partition.
The first order of matter would be to quantify the distance from the current memory partition to our target partition.
To do so, %
we define a $u$-dimensional vector $\delta$ with each element as
\begin{align}
    \delta_i = l_i - p_i \cdot \sum_j l_j, \label{eq:precision}%
\end{align}
where $l_i$ is the number of examples of class $i$ in the memory and $p_i$ is the partition ratio from Eq.~\ref{eq:partition}.
Note that we multiply $p_i$ by $\sum_j l_j$ rather than the memory size $m$, due to the multiple labels on each datapoint.

In order to fulfill our objective (\ie achieve the target partitions), we greedily select and remove the sample that best satisfies the following two desiderata: 
(i) include the classes that occupy more memory than their quota, \ie $\delta_i > 0$ and (ii) exclude the classes that under-occupy or already satisfy the target, $\delta_i \le 0$.

To this end, we devise a two-stage candidate selection process.
For desideratum (i), we define a set of candidate sample $\mathcal Y \subset \mathcal M$ as follows.
Among the classes with $\delta_i > 0$, we randomly sample a class with a probability of softmax$(\delta_i)$.
This sampling is highly biased toward the class with the maximum $\delta_i$ value (\ie the class to be reduced the most).
We found this to be more robust in practice than considering multiple classes with $\delta_i > 0$.
Then, $\mathcal Y$ contains all samples labeled with this selected class in the memory.
For desideratum (ii), we define a $u$-dimensional indicator vector $q$
where $q_i = 0$ if $\delta_i > 0$ and $q_i = 1$ otherwise.
That is, $q_i$ indicates the classes that do not over-occupy the memory. 
Finally, the set of candidate samples $\mathcal{K}$ is obtained by %
\begin{flalign}
    \mathcal{K} = \{n^* | n^* = \argmax_{n \in \mathcal{Y}} (\neg y^n \cdot q) \}, \label{eq:K}
  \end{flalign}
\noindent where $\neg  y^n$ is the negation of the multi-hot label vector of sample $n$ (\ie $0\rightarrow 1$ and $1 \rightarrow 0$).
That is, $\mathcal K$ is a subset of $\mathcal Y$ that does not contain sample(s) for the under-occupied (or already-satisfied) classes as possible.
$\mathcal K$ may include multiple samples while the samples with fewer labels are more likely to be selected.

Finally, amongst $\mathcal{K}$, we select example $z$ to be removed as it is the one that advances the memory towards the target partition the most:
\begin{equation}
    \begin{aligned}
         z = \argmin_{k \in \mathcal{K}} \sum_{i\in\{1,..,u\}} \left\lvert \mathcal{C}_{ki} - p_i \cdot \sum_{l\in\{1,..,u\}}\mathcal{C}_{kl} \right\rvert , %
        \text{where } \mathcal{C}_{ki} = \sum_{n \in \mathcal M \backslash k} y_i^n.  \label{eq:sample_out}
    \end{aligned}
\end{equation}
$\mathcal{C}_{ki}$ is the current number of class $i$ in the memory after the removal of sample $k$ from memory $\mathcal M$, $p_i$ is the partition ratio of class $i$ from Eq.~\ref{eq:partition},
and $y_i^n$ is a binary value for class $i$ of the label vector of sample $n$.
Eq.~\ref{eq:sample_out} finds the sample $z$ that minimizes the distance (defined in  Eq.~\ref{eq:precision}) towards the target partition before and after the removal of sample $k$.

\begin{algorithm}[tb]
    \caption{Partitioning Reservoir Sampling Pseudo-code}
   \label{alg:prs}
   \begin{multicols}{2}
       \begin{algorithmic}[1]
    \REQUIRE (i) data $(x_t, y_t), ... , (x_T, y_T)$, (ii) power param $\rho$, (iii) memory size $m$.
    \STATE $\mathcal{M} = \{\}$ \text{ // memory}
    \STATE $\psi = \O$ \text{ // running statistics}
    \STATE $u = 0$ \text{ // number of unique classes}
   \FOR{$t=1$ {\bfseries to} $T$}
   \STATE $update(\psi)$ \text{// update running stats}
   \IF{$t \le |m|$}
   \STATE \text{// fill memory}
   \STATE $\mathcal{M}_u \leftarrow \mathcal{M}\{y_t\}$ \text{ // sub memory}
   \STATE $\mathcal{M}_u \leftarrow \{x_t, y_t\} \cup \mathcal{M}_u$%
   \ELSE
   \STATE \text{// Partition }
   \STATE $Partitioning(\mathcal{M},\psi,q)$ \text{ // Eq.~\ref{eq:partition}}
   \STATE \text{// Maintenance}
   \STATE $sample\_in(\mathcal{M}_u, y_{t,u}, \psi)$ \text{// Eq.~\ref{eq:sample-in}}
   \IF{\text{sample-in success}}
   \STATE $sample\_out(\mathcal{M}, \psi, q)$ \text{// Eq.~\ref{eq:sample_out}}
   \ENDIF
   \ENDIF
   \ENDFOR
   \end{algorithmic}
   \end{multicols}
\end{algorithm}

\section{Related Work}
There have been three main branches in continual learning, which are regularization, expansion and replay methods.
Here we focus on the replay-based approaches and present a more comprehensive survey in the Appendix.

\textbf{Replay-based approaches}. They explicitly maintain a fixed-sized memory in the form of generative weights or explicit data to rehearse it back to the model during training. %
Many recent works~\cite{isele18aaai,brahma18subset,rolnick19,lopez17,chaudhry19iclr,chaudhry19arxiv,riemer19iclr} employ a memory that reserves the data samples of prior classes in an offline setting.
For example, GEM~\cite{lopez17} uses the memory to constrain the gradient direction that prevents forgetting, and this idea becomes more efficient in AGEM~\cite{chaudhry19iclr}.
Chaudhry \etal\cite{chaudhry19arxiv} explore tiny episodic memory, which shows improved overall performance when training repetitively from only a few examples. %
Riemer \etal\cite{riemer19iclr} introduce a method that combines rehearsal with meta-learning to find the right balance between transfer and interference.
Since our approach uses no prior knowledge other than the given input stream, it is orthogonally integrable with many aforementioned methods.

\textbf{Online Sequential Learning}.
Recently, there have been some approaches to \textit{online} continual learning where each training sample is seen only once.
\textit{ExStream}~\cite{hayes19memory} is an online stream clustering reservoir method, but it requires prior knowledge about the number of classes to pre-allocate the memory.
As new samples enter, the sub-memory is filled based on a distance measure and merged in the feature space when the memory is full.
GSS~\cite{aljundi19gradient} may be one of the most similar works to ours. %
It formulates the sample selection as a constraint reduction problem,
intending to select a fixed subset of constraints that best approximate the feasible region. %
They perform miniaturized MNIST experiments with different task sizes (\eg 2000 instances for one task and 200 for the others). %
However, this setting is difficult to represent practical long-tailed or imbalanced problems, since only a single task is much larger than the other same-sized tasks. %

\textbf{Multi-label Classification}. 
There have been many works handling the vital problem of multi-label classification~\cite{makadia08eccv,guillaumin09iccv}.
Recently, recurrent approaches~\cite{wang16cvpr,wang17iccv} and attention-based methods~\cite{zhu17cvpr,guo19cvpr}  are proposed to correlate the labels during predictions.
Wei \etal\cite{chen19cvpr} employ the prior task knowledge to perform graph-based learning that aids the correlation representation of multiple labels.
While all the works in the past have focused on the offline multi-label classification, we take on its  online task-free continual learning problem.
Moreover, our approach is orthogonally applicable to these methods as we select some of them as the base model in our experiments.

\begin{figure}[t!]
    \begin{center}
    \includegraphics[width=\textwidth]{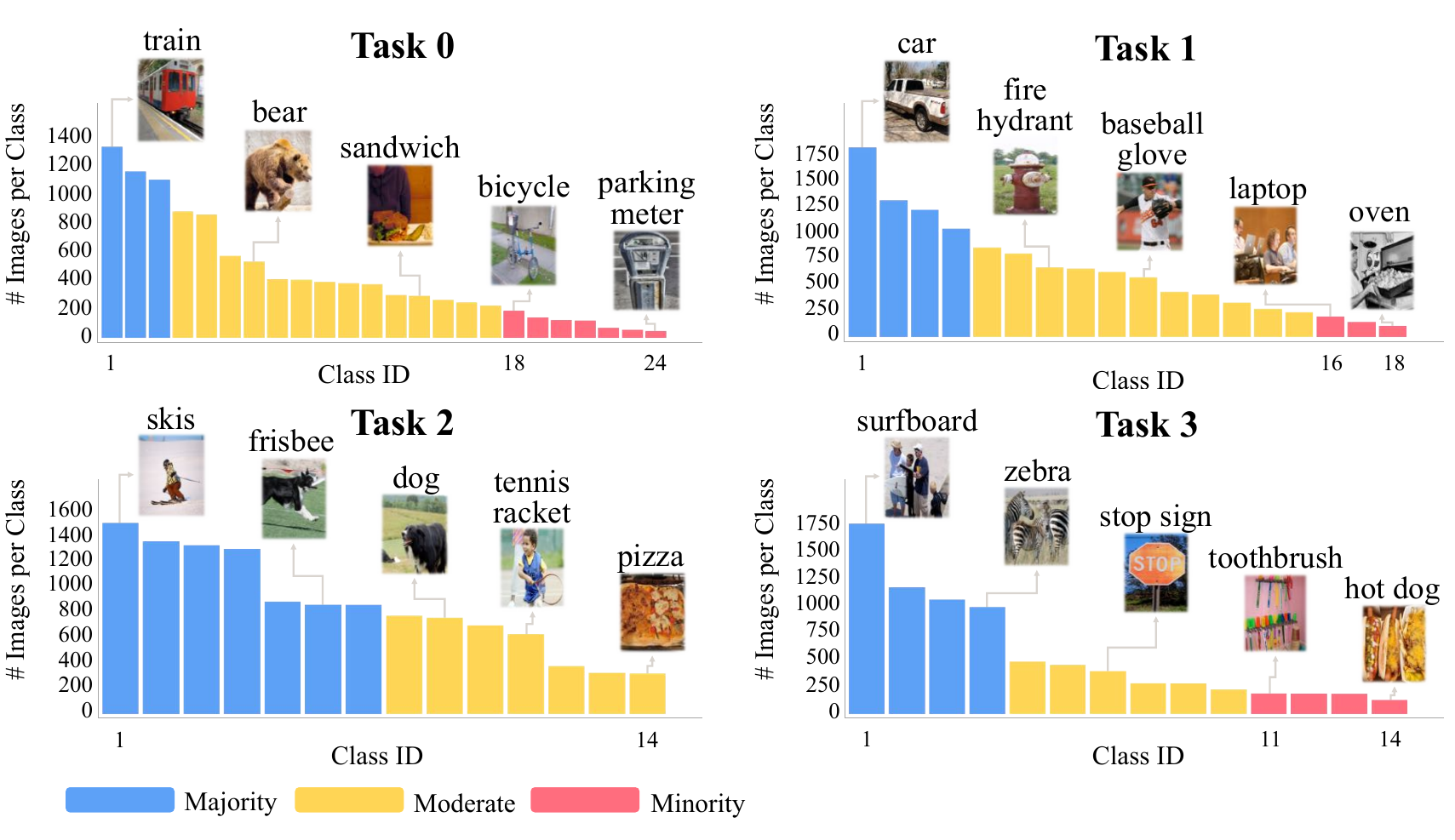}
    \caption{\textbf{Statistics of \textit{COCOseq} dataset} consisting of four tasks.}
    \label{fig:coco_sequence}
    \end{center}
\end{figure}

\section{The Multi-label Sequential Datasets}
\label{sec:dataset}
To study the proposed problem, we transform two multi-label classification datasets into their continual versions: \textit{COCOseq} and \textit{NUS-WIDEseq}.
It is a non-trivial mission since the data must be split into tasks with exclusive class labels where each datapoint is associated with multiple labels.

\subsection{The \textit{COCOseq}}
There have been two previous works that curate the MSCOCO dataset~\cite{lin14eccv} for continual learning.
Shmelkov \etal\cite{shmelkov17iccv} select 20 out of 80 classes to create 2 tasks, each with 10 classes by grouping them based on alphabetical ordering.
They also create an incremental version with 25 classes, where 15 classes are used for the model to obtain the base knowledge via normal batch training, and the other 10 classes are sequentially learned one at a time.
These 10 classes are selected so that each image has only a \textit{single} class label.
Nguyen \etal\cite{nguyen19arxiv} tailor MSCOCO for continual learning of captioning.
They use 24 out of 80 classes and discarded all the images that contain multiple class labels.
Similar to \cite{shmelkov17iccv}, they create 2 tasks where one task has 5 classes, and the other has 19 classes.
Also, a sequential version is made using the 19 classes  for base knowledge learning and the remaining 5 for incremental learning.

Different from previous works, we curate 4 tasks %
with multi-label images.
To accurately measure the training performance on the intra-task imbalance, we make sure that the test set is \textit{balanced};
the test set size per class is identical even though its training set size is imbalanced and long-tailed.
This is a common practice in imbalanced classification benchmarks, including \cite{wang17nips} that uses 40 test images per class in the SUN-LT dataset and the largest OLTR benchmark \cite{liu19large} that assigns 50 and 100 balanced test images per class for ImageNet-LT and Places-LT dataset, respectively.
While referring to the Appendix for more details of dataset construction, we build a 4-way split MSCOCO dataset called \textit{COCOseq} (Figure~\ref{fig:coco_sequence}), consisting of 70 object classes with 35072 training and 6346 test data.
The test set contains one-hundred images per class. Note that $6346 \neq 70 \times 100$ due to the multi-label property.

To the best of our knowledge, there is no strict consensus to divide the majority and minority classes in long-tailed datasets.
For instance, Liu \etal\cite{liu19large} define the classes with more than 100 training examples as many-shot, 20-100 as medium-shot, and less than 20 as few-shot classes.
Other works such as \cite{han18emnlp} and \cite{zhang19naacl} define the minority classes as less than 100 or 200 samples in the training set, respectively.
Accordingly, we define classes with less than 200 training examples as the minority classes, 200-900 as moderate and $>$900 as the majority.

\subsection{The \textit{NUS-WIDEseq}}
We further curate a sequential dataset from NUS-WIDE~\cite{chua09acm}, containing 6 mutually exclusive and increasingly difficult tasks.
Its novelty lies in having both inter- and intra-task imbalance; the skewness exists not only within each task but amongst the tasks as well.
More details can be found in the Appendix.

NUS-WIDE~\cite{chua09acm} is a raw web-crawled multi-label image dataset.
It provides a human-annotated version with 150531 images of 81 labels\footnote{The original number of images is 210832, but many URLs are no longer available.}.
However, the dataset by nature exhibits a very severe long-tail property. For instance, MSCOCO's top 20\% classes are responsible for 50.7\% of the total data,
while NUS-WIDE's top 20\% surmount to 76.3\% of the whole data.
Since the original test set is highly long-tailed, we balance it for more accurate evaluation as done for \textit{COCOseq}. %
Finally, \textit{NUS-WIDEseq} contains 49 classes with 48724 training and 2367 test data with 50 samples per class.

\section{Experiments}
\label{sec:experiments}
In our evaluation, we explore how effective our PRS  is for both inter- and intra-task imbalances in task-free multi-label continual learning tasks compared to the state-of-the-art models.
We also analyze the importance of a balanced memory from many aspects. 
The task that we solve is mostly close to but more difficult than the scenario of \textit{class-incremental learning} \cite{vandeven19}
in that the task label is not available at training as well as at test time.

\subsection{Experimental Design}
Previous continual learning research has shown a high amount of disparity in evaluation.
As we are the first to explore multi-label continual learning,
we explicitly ground our experimental setting based on \cite{farquhar19,aljundi19thesis,vandeven19} as follows:
\begin{enumerate}[label=$\bullet$]
    \item \textit{Cross-task resemblance}: Consecutive tasks in COCOseq and NUS-WIDEseq are partly correlated to contain neighboring domain concepts. 
    \item \textit{Shared output heads}: Since we solve multi-label classification without task labels, the level of difficulty of our task is comparable to using a shared output head for single-label classification.
    \item \textit{No test-time task labels}: Our approach does not require explicit task labels during both training and test phase, often coined as \textit{task-free continual learning} in \cite{aljundi19gradient,lee20iclr}.
    \item \textit{More than two tasks}: COCOseq and NUS-WIDEseq contain four and six tasks, respectively.
    \item \textit{Online learning}: The algorithm learns from a continuous stream of data without a separate offline batch training stage such as \cite{aljundi19gradient,lee20iclr,hayes19memory}.
\end{enumerate}

\begin{table*}[t!]
    \centering
    \scriptsize
    \setlength{\tabcolsep}{2pt}
    \caption{\textbf{Results on \textit{COCOseq} and \textit{NUS-WIDEseq}}. We report accuracy metrics for multi-label classification  after the whole data stream is seen once. Similar to \cite{liu19large},
    the majority, moderate and minority are distinguished to accurately assess the long-tail performances.
    The memory size is fixed at 2000, with \{0,3,1,2\} task schedule for \textit{COCOseq} and \{3,1,0,5,4,2\} for \textit{NUS-WIDEseq}.
    The results are the means of five experiments except those of GSS-Greedy~\cite{aljundi19gradient} which are the mean of three due to its computational complexity. 
    The best and the second best methods are respectively marked in \textcolor{red}{red} and \textcolor{blue}{blue} fonts, excluding the MULTITASK that is offline trained as the upper-bound.
   FORGET refers to the normalized forgetting measure of~\cite{chaudhry18eccv}.}
    \label{tab:coco_results}
    \begin{sc}
        \begin{tabular}{lccc|ccc|ccc||ccc}
            \toprule
            &\multicolumn{3}{c|}{majority} & \multicolumn{3}{c|}{moderate}& \multicolumn{3}{c||}{minority}& \multicolumn{3}{c}{\textbf{Overall}}\\
            \textbf{\textit{COCOseq}}      & C-F1 & O-F1 & mAP  & C-F1 & O-F1 & mAP  & C-F1 & O-F1 &  mAP & C-F1 &  O-F1 &  mAP \\
            \midrule
            Multitask~\cite{caruaca97}       & 72.9 & 70.9 & 77.3 &  53.2  &  51.4 & 55.0 & 12.7 &  13.6 & 24.2 & 51.2 & 52.1  & 53.9 \\
            \specialrule{0.1pt}{1pt}{1pt}
            \specialrule{0.1pt}{1pt}{1pt}
            Finetune  & 18.5 & 27.9 & 29.8 & 6.7  & 16.7 & 14.1 & 0.0  & 0.0  & 5.2  & 8.5  & 18.4  & 16.4 \\
            \;forget   & 100.0& 100.0& 65.8 & 100.0& 100.0& 73.5 & 100.0& 100.0& 67.4 & 100.0& 100.0 & 70.1 \\

            \specialrule{0.1pt}{1pt}{1pt}
            EWC~\cite{kirkpatrick17ewc} & 60.0 & 53.4 & 64.1 & 37.3 & 38.1 & 47.5 &  7.5 & 8.2  & 21.5 & 38.9 & 40.0  & 46.6 \\
            \;forget   & 24.2 & 24.0 & 0.8  & 35.1 & 33.9 & 3.0  & 56.3 & 56.1 & 9.0  & 32.8 & 32.0  & 3.2 \\

            \specialrule{0.1pt}{1pt}{1pt}
            CRS~\cite{vitter85} & \textcolor{red}{67.0} & \textcolor{red}{62.5} & \textcolor{red}{67.9} & 47.8 & 45.2 & 50.4 & 14.5 & 15.6 & 26.9 & 47.5 & \textcolor{blue}{46.6}  & 50.2 \\
            \;forget   & 15.0 & 13.6 &  8.9 & 32.8 & 32.0 & 15.6 & 55.58& 54.92& 23.2 & 32.2 & 30.1  & 15.3 \\

            \specialrule{0.1pt}{1pt}{1pt}
            GSS~\cite{aljundi19gradient} & 59.3 & 56.7 & 59.6 & 44.9 & 43.0 & 46.0 & 10.5 & 11.0 & 18.6 & 42.8 & 42.7 & 44.0 \\
            \;forget   & 20.2 & 18.8 & 10.3 & 36.4 & 35.3 & 13.6 & 67.4 & 68.4 & 26.1 & 35.1 & 35.3 & 13.6 \\
            \specialrule{0.1pt}{1pt}{1pt}
            ExStream~\cite{hayes19memory} & 58.8 & 52.0 & 62.5 & \textcolor{blue}{49.2} & \textcolor{blue}{47.3} & \textcolor{blue}{52.7} & \textcolor{blue}{26.4} & \textcolor{blue}{26.6} & \textcolor{blue}{36.6} & \textcolor{blue}{47.8} & 43.9 & \textcolor{blue}{51.1} \\
            \;forget & 41.8 & 40.2 & 17.7 & 33.9 & 32.9 & 14.7 & 47.0 & 33.4 & 15.5 & 40.5 & 39.6 & 16.2 \\
            \specialrule{0.4pt}{1pt}{1pt}
            PRS(ours) & \textcolor{blue}{65.4} & \textcolor{blue}{59.3} & \textcolor{blue}{67.5} & \textcolor{red}{52.5} & \textcolor{red}{49.7} & \textcolor{red}{55.2} & \textcolor{red}{34.5} & \textcolor{red}{34.6} & \textcolor{red}{39.7} & \textcolor{red}{53.2} & \textcolor{red}{50.3} & \textcolor{red}{55.3} \\
            \;forget & 22.0 & 21.7 & 8.5 & 27.2 & 26.8 & 11.5 & 26.2 & 26.3 & 10.5 & 25.6 & 25.2 & 10.2 \\
            \bottomrule
            \\
            \toprule
                     &\multicolumn{3}{c|}{majority} & \multicolumn{3}{c|}{moderate}& \multicolumn{3}{c||}{minority}& \multicolumn{3}{c}{\textbf{Overall}}\\
            \textbf{\textit{NUS-WIDEseq}}   & C-F1 & O-F1 & mAP  & C-F1 & O-F1 & mAP  & C-F1 & O-F1 &  mAP & C-F1 &  O-F1 &  mAP \\
            \midrule
            Multitask~\cite{caruaca97}      & 33.7 & 30.8 & 32.8 &  29.3  &  28.7 & 28.9 &  9.7 &  11.8 & 25.8 & 24.6 & 24.9  & 28.4 \\

            \specialrule{0.1pt}{1pt}{1pt}
            \specialrule{0.1pt}{1pt}{1pt}
            Finetune  & 0.6 & 4.6 & 4.1 & 2.3 & 2.8 & 6.0 & 5.2 & 7.4 & 9.4 & 4.2 & 5.1 & 7.1 \\
            \;forget & 100.0 & 100.0 & 47.3 & 100.0 & 100.0 & 39.3 & 100.0 & 100.0 & 44.6 & 100.0 & 100.0 & 44.4 \\

            \specialrule{0.1pt}{1pt}{1pt}
            EWC~\cite{kirkpatrick17ewc} & 15.7 & 9.9 & 15.8 & 16.3 & 12.6 & 19.4 & 12.3 & 13.9 & 24.1 & 17.1 & 11.4 & 20.7 \\
            \;forget & 18.4 & 15.4 & 7.8 & 64.6 & 63.7 & 7.3 & 63.4 & 63.4 & 4.8 & 36.4 & 31.5 & 7.3 \\

            \specialrule{0.1pt}{1pt}{1pt}
            CRS~\cite{vitter85} & \textcolor{red}{28.4} & \textcolor{red}{17.8} & \textcolor{red}{21.9} & 13.6 & 14.2 & \textcolor{blue}{18.5} & 10.4 & 11.8 & 20.6 & 16.8 & 15.0 & 20.1 \\
            \;forget & 33.6 & 29.2 & 14.7 & 67.8 & 66.7 & 18.1 & 96.5 & 96.2 & 20.3 & 61.5 & 57.8 & 18.7 \\

            \specialrule{0.1pt}{1pt}{1pt}
            GSS~\cite{aljundi19gradient} & 24.6 & 13.5 & 19.0 & \textcolor{blue}{14.8} & \textcolor{blue}{15.5} & 17.9 & 15.9 & 17.6 & 24.5 & 17.9 & 15.3 & 20.9 \\
            \;forget & 46.8 & 43.9 & 16.6 & 59.6 & 58.3 & 11.7 & 82.8 & 82.0 & 18.6 & 54.8 & 49.8 & 13.0 \\

            \specialrule{0.1pt}{1pt}{1pt}
            ExStream~\cite{hayes19memory} & 15.6 & 9.2 & 15.3 & 12.4 & 12.8 & 17.6 & \textcolor{blue}{24.6} & \textcolor{blue}{24.1} & \textcolor{blue}{26.7} & \textcolor{blue}{18.7} & \textcolor{blue}{16.0} & \textcolor{blue}{21.0} \\
            \;forget & 80.7 & 77.6 & 24.0 & 81.0 & 80.6 & 23.3 & 77.2 & 76.7 & 21.8 & 81.0 & 79.3 & 23.4 \\

            \specialrule{0.4pt}{1pt}{1pt}
            PRS(ours) & \textcolor{blue}{26.7} & \textcolor{blue}{17.9} & \textcolor{blue}{21.2} & \textcolor{red}{19.2} & \textcolor{red}{19.3} & \textcolor{red}{21.5} & \textcolor{red}{27.5} & \textcolor{red}{26.8} & \textcolor{red}{31.0} & \textcolor{red}{24.8} & \textcolor{red}{21.7} & \textcolor{red}{25.5} \\
            \;forget & 45.8 & 43.0 & 15.7 & 59.0 & 58.4 & 13.4 & 60.6 & 60.3 & 15.5 & 55.3 & 53.5 & 13.9 \\
            \bottomrule
        \end{tabular}
    \end{sc}
\end{table*}

\textbf{Base Models}.
In recent multi-label image classification~\cite{wang16cvpr,wang17iccv,liu19acm,chen19cvpr,guo19cvpr},
it is a common practice to fine-tune a pre-trained model to a target dataset.
We thus employ ResNet101~\cite{he16cvpr} pre-trained on ImageNet~\cite{deng09imagenet} as our base classifier.
Additionally, we test two multi-label classification approaches that \textit{do not} require
any prior information about the input to train:
Recurrent Attention (RNN-Attention)~\cite{wang17iccv}
and the more recent Attention Consistency (AC) algorithm~\cite{guo19cvpr}.
Due to its superior performance, we choose ResNet101 as the base model for the experiments in the main draft.
We report the results of RNN-Attention and AC methods in the Appendix.

\textbf{Evaluation Metrics}.
Following the convention of multi-label classification~\cite{wang16cvpr,zhu17cvpr,ge18cvpr},
we report the average overall F1 (O-F1), per-class F1 (C-F1) as well as the mAP.
Additionally, we include the forgetting metric~\cite{chaudhry18eccv} to quantify the effectiveness of continual learning techniques.
However, since this is a self-relative metric on the best past and present performance of the method,
comparisons between different methods could be misleading
(\eg if a model performs poorly throughout training, small forgetting metric values can be observed as it has little information to forget from the beginning).
It is the reason for the absence of color for the best models with respect to this metric in the tables.
In the Appendix, we also report the overall precision (O-P), recall (O-R), per-class precision (C-P) and recall (C-R) metrics.

\textbf{Baselines.}
We compare our approach with six baselines including four state-of-the-art continual learning methods: EWC~\cite{kirkpatrick17ewc}, CRS~\cite{vitter85}, GSS-Greedy~\cite{aljundi19gradient} and ExStream~\cite{hayes19memory}.
In addition, the Multitask~\cite{caruaca97} can be regarded as an upper-bound performance as it is learned offline with minibatch training for a single epoch.
The Finetune performs online training without any continual learning technique, and thus it can be regarded as a lower-bound performance.
For training EWC, we fix the ResNet up to the penultimate layer in order to obtain sensible results; otherwise, it works poorly.
More details for baselines are presented in the Appendix.

We use a fixed online input batch size of $10$ and a replay-batch size of 10 in accordance with \cite{aljundi19gradient}.
We use Adam~\cite{kingma15} optimizer with $\beta_1=0.9$, $\beta_2=0.999$ and $\epsilon=1e-4$, and finetune all the layers unless stated otherwise.
Furthermore, we set $\rho$ between the range of [-0.2, 0.2],
and fix the memory size to 2000 (as done in \cite{maltoni19elsevier}), which are $5.7\%$ and $4.7\%$ of the overall training data for \textit{COCOseq} and \textit{NUS-WIDEseq}, respectively.

\begin{table*}[t!]
    \begin{minipage}{0.52\textwidth}
    \centering
    \scriptsize
    \setlength{\tabcolsep}{2pt}
    \caption{\textbf{Results according to memory sizes and schedule permutations} on \textit{COCOseq}. We fix the memory size of 2000 for schedule experiments and the schedule of \{0,3,1,2\}  for memory experiments. 
    Refer to Table~\ref{tab:coco_results} for the nomenclatures.}
    \label{tab:reservoir_size_results}
    \begin{sc}
        \begin{tabular}{lccc|ccc}
       \toprule
       & \multicolumn{3}{c|}{\textbf{Overall}}& \multicolumn{3}{c}{\textbf{Overall}}\\
       \textbf{\textit{COCOseq}}        & C-F1 & O-F1 &  mAP &  C-F1 &  O-F1 &  mAP \\
       \midrule
       & \multicolumn{3}{c|}{Schedule: 0, 1, 3, 2}& \multicolumn{3}{c}{memory: $1000$}\\
       \specialrule{0.1pt}{1pt}{1pt}

       CRS\cite{vitter85} & \textcolor{blue}{49.2}  & \textcolor{blue}{46.9} & \textcolor{blue}{50.8} & \textcolor{blue}{44.2} & \textcolor{blue}{41.6}  & \textcolor{blue}{47.1} \\
       GSS\cite{aljundi19gradient} & 42.1  & 41.4  & 44.0 & 40.6 & 39.0  & 42.1 \\
       ExStream\cite{hayes19memory} & 47.3 & 42.9 & 50.5 & 41.6 & 37.6 & 47.0 \\
       \specialrule{0.2pt}{1pt}{1pt}
       PRS(ours) & \textcolor{red}{52.6} & \textcolor{red}{50.5} & \textcolor{red}{55.2} & \textcolor{red}{47.4} & \textcolor{red}{43.4} & \textcolor{red}{51.2} \\
       \specialrule{0.4pt}{1pt}{1pt}

       & \multicolumn{3}{c|}{Schedule: 2, 3, 0, 1}& \multicolumn{3}{c}{memory: $2000$}\\
       \specialrule{0.1pt}{1pt}{1pt}
       CRS\cite{vitter85} & \textcolor{blue}{45.4} & \textcolor{blue}{44.3} & \textcolor{blue}{48.2} & 47.5 & \textcolor{blue}{46.6}  & 50.2 \\
       GSS\cite{aljundi19gradient} & 33.5 & 33.9 & 38.5 & 43.2 & 43.0  & 44.4 \\
       ExStream\cite{hayes19memory} & 41.6 & 35.2 & 45.7 & \textcolor{blue}{47.8} & 43.9 & \textcolor{blue}{51.1} \\
       \specialrule{0.2pt}{1pt}{1pt}
       PRS(ours) & \textcolor{red}{50.4} & \textcolor{red}{47.7} & \textcolor{red}{53.1} & \textcolor{red}{53.2} & \textcolor{red}{50.3} & \textcolor{red}{55.3} \\

       \specialrule{0.4pt}{1pt}{1pt}
       & \multicolumn{3}{c|}{Schedule: 3, 1, 0, 2}& \multicolumn{3}{c}{memory: $3000$}\\
       \specialrule{0.1pt}{1pt}{1pt}
       CRS\cite{vitter85} & \textcolor{blue}{47.7} & \textcolor{blue}{45.9} & \textcolor{blue}{49.7} & 49.4 & \textcolor{blue}{48.6}  & 51.1 \\
       GSS\cite{aljundi19gradient} & 37.8  & 38.8 & 40.2 & 42.2 & 43.0  & 44.3 \\
       ExStream\cite{hayes19memory} & 45.4 & 41.8 & 49.2 & \textcolor{blue}{49.7} & 46.8 & \textcolor{blue}{52.2} \\
       \specialrule{0.2pt}{1pt}{1pt}
       PRS(ours) & \textcolor{red}{51.4} & \textcolor{red}{48.9} & \textcolor{red}{54.1} & \textcolor{red}{54.9} & \textcolor{red}{53.4} & \textcolor{red}{56.7} \\
       \bottomrule
   \end{tabular}
    \end{sc}
\end{minipage}%
\hspace{0.5cm}%
\begin{minipage}{0.43\textwidth}
    \includegraphics[width=\textwidth]{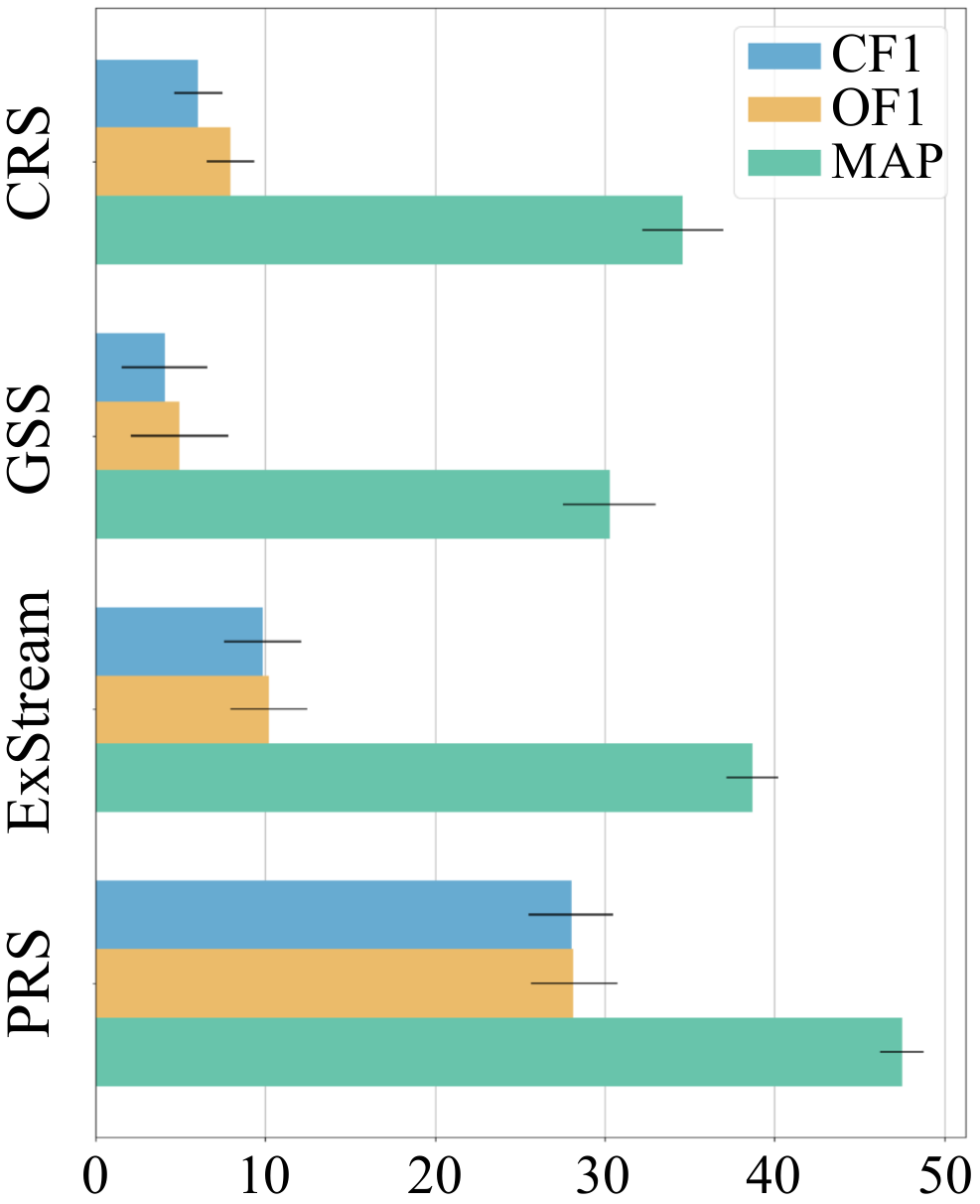}
    \captionof{figure}{\textbf{Inter-task imbalance analysis} on \textit{NUS-WIDEseq}. We compare the performance for the smallest Task 3, for which our PRS robustly outperforms all the baselines.}
    \label{fig:inter_task_imbal}
\end{minipage}
\end{table*}

\subsection{Results}

Table~\ref{tab:coco_results} compares continual learning performance between our PRS method and baselines on \textit{COCOseq} and \textit{NUS-WIDEseq}.
In all comparable metrics of C-F1, O-F1 and mAP,
PRS outperforms CRS~\cite{riemer19iclr}, GSS~\cite{aljundi19gradient} and even ExStream~\cite{hayes19memory} that uses prior task information to pre-allocate the memory.

\textbf{Schedule and Memory Permutations.}
Table~\ref{tab:reservoir_size_results} compares the robustness of PRS through random permutations of task schedule as well as different memory sizes.
Interestingly, the performances of CRS, GSS and ExStream fluctuates depending on the permuted schedule, while our PRS is comparatively robust thanks to the balanced emphasis on all the learned classes.
Moreover, PRS outperforms all the baselines with multiple memory sizes of $1000, 2000, 3000$. %

\textbf{Intra- and Inter-task Imbalance.} %
Table~\ref{tab:coco_results} shows that PRS is competitive on the majority classes (\eg marked in blue as the runner-up) and performs the best on both moderate and minority classes, showing its compelling robustness for the intra-task imbalances.
Furthermore, Figure~\ref{fig:inter_task_imbal} validates the robustness of PRS in the inter-task imbalance setting. %
As shown in Fig. 1 of the Appendix, tasks of NUS-WIDEseq are \textit{imbalanced} in that the smallest Task 3 is $9.6$ times smaller than that of the largest Task 1.
Figure~\ref{fig:inter_task_imbal} compares the performances of all methods for the minority Task 3, for which 
PRS performs overwhelmingly better than the other baselines in all the metrics.

\begin{figure}[!tbp]
  \centering
  \begin{minipage}[b]{0.49\columnwidth}
    \includegraphics[width=\columnwidth]{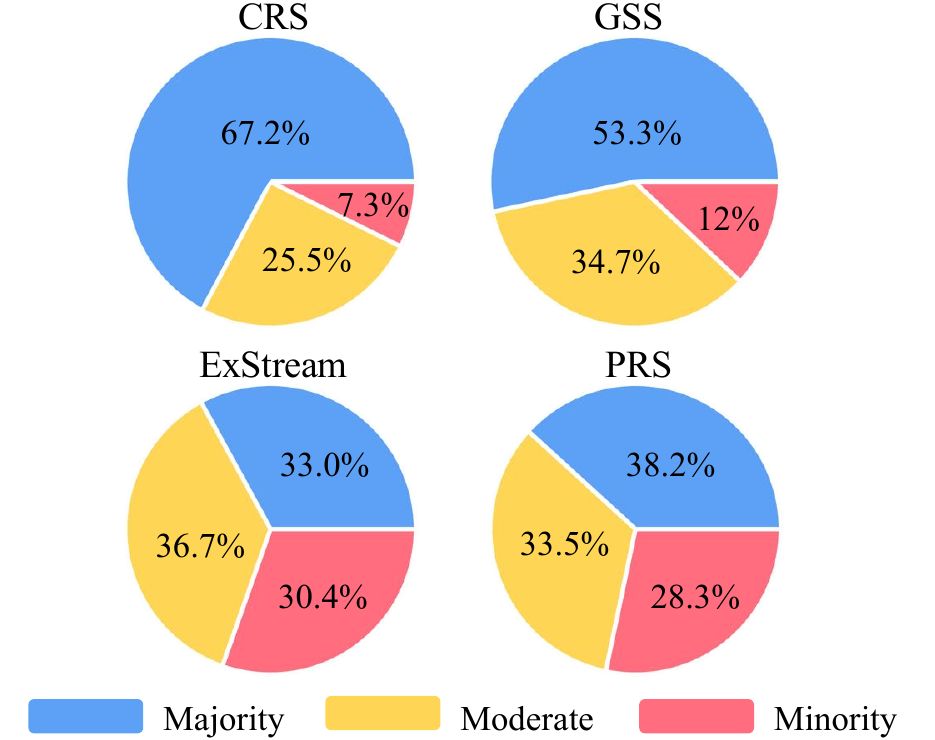}
    \caption{The resulting memory distribution of the \textit{COCOseq} tests in Table \ref{tab:coco_results}. ExStream \cite{hayes19memory} is the only \textit{task-aware} method that knows the task distribution beforehand.}
    \label{fig:buffer_distribution}
  \end{minipage}
  \hfill
  \begin{minipage}[b]{0.49\columnwidth}
    \includegraphics[width=\columnwidth]{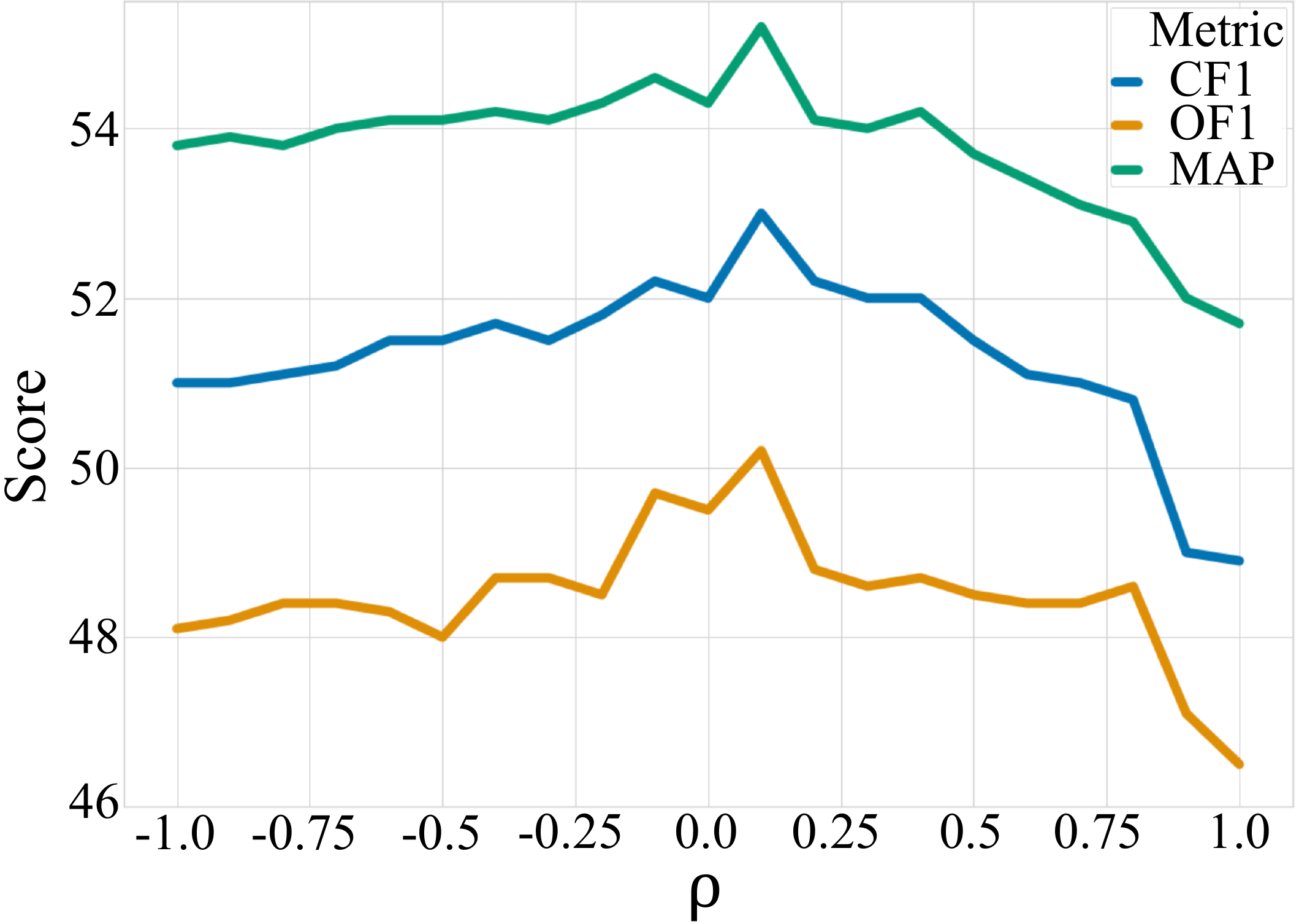}
    \caption{Performance of PRS with different $\rho$ in the range of [-1, +1] on \textit{COCOseq}.
    All results are the averages of 5 different random seeds.}
    \label{fig:q_dist}
  \end{minipage}
\end{figure}

\textbf{Memory Distribution After Training.}
Figure~\ref{fig:buffer_distribution} compares the normalized memory distribution of the experiments in Table \ref{tab:coco_results}.
CRS dominantly uses the memory for the majority classes while reserving only a small portion for the minority.
This explains why CRS may perform better than PRS for the majority in Table \ref{tab:coco_results}, while 
sacrificing performance largely for the moderate and minority classes.
On the other hand, GSS saves much more samples for the moderate classes relative to CRS, but still fails to maintain a sufficient number of samples for the minority.
Note that Exstream balances the memory using \textit{prior task information}.
However, due to its clustering scheme via feature merging to maintain the memory, it is difficult to obtain representative clusters, especially when handling complex datasets with multi-labels.
Importantly, PRS can balance the memory for all classes without any auxiliary task information.

\textbf{Power of Allocation $\rho$.}
Figure~\ref{fig:q_dist} shows the performance variation according to different  $\rho$. It confirms that a balance of memory is vital for the performance even when the input stream is highly imbalanced.
Notice, as $\rho$ moves away from the vicinity of balance, the performance gradually declines in all metrics. %

\begin{figure}[t!]
    \begin{center}
    \includegraphics[width=\columnwidth]{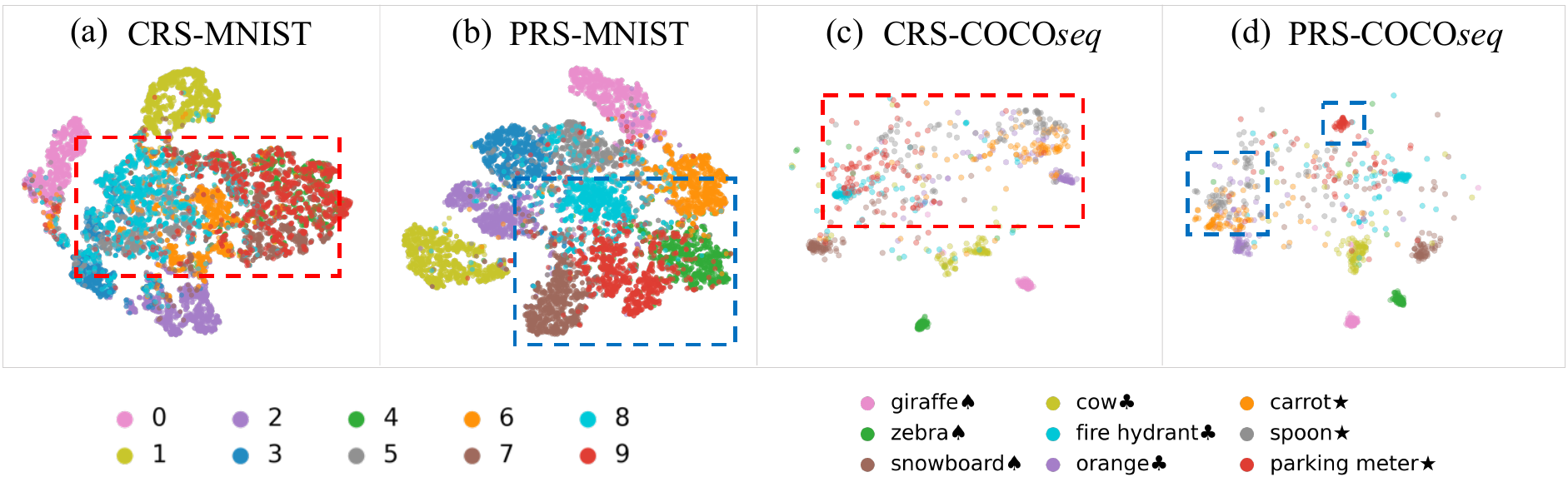}
    \caption{\textbf{t-SNE feature projection} of CRS and PRS trained features for test samples of (a)--(b) MNIST and (c)--(d) COCOseq.
        We use the penultimate features of a 2-layer feedforward for MNIST and ResNet101 for COCOseq.
        As with the single-label experiments in the Appendix, we curate a %
        sequential MNIST that follows a Pareto distribution~\cite{reed01} with a power value $\alpha$=0.6, which becomes increasingly long-tailed from 0 to 9. 
        For COCOseq, %
        we use the symbols of \{Minority: $\bigstar$, Moderate: $\clubsuit$, Majority: $\spadesuit$\}.
        We emphasize that PRS represents the minority classes (in the blue box) much more discriminatively than the corresponding class features (in the red box) for CRS.}
    \label{fig:t_sne}
    \end{center}
\end{figure}

\textbf{Feature Analysis}
Figure~\ref{fig:t_sne} shows the feature projections of CRS and PRS for test samples of MNIST and COCOseq using t-SNE~\cite{vanDerMaaten2008}.
PRS can represent the minority classes more discriminatively than CRS on both single-label MNIST and multi-label COCOseq experiments.

In the Appendix, we include more experimental results, including analysis on the memory gradients and performance on single-label classification and many more.

\section{Conclusion}
\label{sec:conclusion}

This work explored a novel problem of multi-label continual learning, which naturally requires the model to learn from imbalanced data streams. %
We contributed two datasets and an effective memory
maintenance algorithm, called \textit{Partitioning Reservoir Sampling} to tackle this new challenge.
Our results showed the importance of maintaining a well-rounded knowledge through balanced replay memory.
As a future direction of research, the ability to learn online while automatically tuning the target partitions would be an exciting avenue to explore.

\smallskip
\noindent \textbf{Acknowledgements.} We express our gratitude for the helpful comments on the manuscript by Soochan Lee, Junsoo Ha and Hyunwoo Kim.
This work was supported by Samsung Advanced Institute of Technology,
Institute of Information \& communications Technology Planning \& Evaluation (IITP) grant (No.2019-0-01082, SW StarLab) and
the international cooperation program by the NRF of Korea (NRF-2018K2A9A2A11080927).
Gunhee Kim is the corresponding author.

\clearpage
\bibliographystyle{splncs04}
\bibliography{eccv2020}

\section{Appendix}
\label{sec:appendix}
In the main manuscript, we revealed the new problem of fatal forgetting in the minority, and curated 2 datasets, \textit{COCOseq} and \textit{NUSWIDEseq}
to study the problem. In addition, we proposed the \textit{Partitioning Reservoir Sampling}, which can effectively maintain the memory towards the desired partition ratios.
In the supplemental material, we enlist the following which may shed further insights:

\begin{enumerate}[label=\Roman*.]
    \item Additional Related Work Sec.~\ref{sec:add_related}
    \item Dataset Curation Details Sec.~\ref{sec:dataset_curation}
    \begin{enumerate}[label=\roman*.]
        \item Imbalanced Multi-label Sequential Dataset Sec.~\ref{sec:multi_label_seq_dataset}
        \item Imbalanced Single-label Sequential Dataset Sec.~\ref{sec:single_label_seq_dataset}
    \end{enumerate}
    \item Experiment Details Sec.~\ref{sec:add_experiments}
    \item Results \& Analysis Sec.~\ref{sec:add_analysis}
    \begin{enumerate}[label=\roman*.]
        \item Single-label results Sec.~\ref{sec:single_label_results}
        \item Gradient Variance analysis Sec.~\ref{sec:grad_var}
    \end{enumerate}

\end{enumerate}
Notably, in the Results \& Analysis section, we report the results for imbalanced single-label experiments in Sec.~\ref{sec:single_label_results}
where PRS also performs superior to the other baselines.

\subsection{Related Work}
\label{sec:add_related}
\subsubsection{Continual Learning.} additional continual learning approaches will be introduced here.

\textbf{Generative Replay Methods.}
The first Generative Replay approach was introduced by \cite{shin17}. The authors proposed a dual-model architecture which consists of a deep generative model
and a task solver (e.g., a classifier). The training data from previously learned tasks are sampled as pseudo-data via a generative model.

Soon after, \cite{kemker18} proposed the FearNet, which uses both a small reservoir as well as a generative network that creates pseudo-samples. During learning, both are
juxtaposed into a replay reservoir to prevent forgetting.
There also exists a host of other generative replay approaches as the direction itself is quite promising~\cite{wu18memgans,lesort19a,lesort19b}.

\textbf{Regularization.}
\label{subsec:regularization}
uses a soft constraint in the form of a penalty on the loss function to regularize the parameters.
It can generally be formulated as below, 
\begin{align}
& \mathcal{\hat{L}}^n(\theta) =\mathcal{L}^n(\theta) + \lambda \sum_i \delta_i h(\theta^n_i - \theta^{n-1}_i)
\end{align}
where $\delta_i$ denotes the importance parameters for each of the weights, $\theta^{n-1}_i$ denotes the learned weights up to the previous task.
We can see that the parameters gradually drift away from the earlier tasks. Hence, it becomes especially vulnerable when there are long sequences of tasks to be solved.
Learning without Forgetting \cite{li16lwf} was an approach in which the knowledge distillation technique was used for the previously fitted
network parameters to be maintained during the learning of a new task. 
Soon after, EWC was proposed by \cite{kirkpatrick17ewc}. It uses a penalty between the parameters 
of previous and new tasks to slow down the forgetting in which a Fisher information matrix approximates the importance of individual parameters.
Along a similar vein, \cite{zenke17} introduced a method to alleviate forgetting via weight importance estimation for the current task at hand.
More recently, \cite{aljundi19selfless} proposed a method in which enforced a sparsity of the learned tasks so that weights can be parsimoniously used to save space
for future tasks while mitigating forgetting.
\textbf{Expansion.}
\label{subsec:expansion}
While the regularization methods can be interpreted as giving a soft constraint, an explicitly hard constraint can be given to the learning process by
freezing the learned tasks and expanding the parameters for the newly learned tasks.
\cite{rusu16} first proposed Progressive Networks to prevent any updates to the parameters trained on the previous tasks by freezing them and allocating a new set of parameters for the incoming
future tasks with lateral connections to the previous parameters.
Neurogenesis Deep Learning (NDL)~\cite{draelos17}  model adds new neural units to the autoencoder to facilitate the addition of new samples along with replay.
\cite{yoon18} extended the concept on the supervised learning setting and proposed the Dynamically Expandable Networks (DEN), which laterally expands the number of trainable
parameters to learn new tasks while using the $L2$ regularization to perform sparse, selective retraining allowing it to decide how many neurons to expand at each layer.

\subsubsection{Imbalance.}
\label{sec:imbalance}
The imbalance problem is an innate attribute of the real world~\cite{zhu14cvpr,van17devil}.
A substantial imbalance leads to models that exhibit a bias towards the majority concept and, at extreme levels, even ignore the minority concept altogether.
In the machine learning community, there exists the data-based and algorithm-based approaches.

\textbf{Data-based Approach.}
The data-based approaches include over-sampling and under-sampling methods.
The motivation is to augment the training distributions in order to decrease the level of imbalance and solve the problem at its root.
The over-sampling methods duplicate the data of minority near the tail distribution.
However, increased training time and a higher risk of over-fitting are some of its well-known downsides. There have been some important techniques to
balance these trade-offs \cite{chawla02smote,maciejewski11local,shen16relay}.

With the surge of better generative models, there has been an exploration into the generation of synthetic data of the minority concepts \cite{chawla02smote,maciejewski11local,douzas18effective}.
It is a difficult problem since the minority concepts lack the number of data to model its inherent properties accurately, but it is a positive direction to pursue.

Under-sampling techniques voluntarily hold out on information by discarding some of the majority classes \cite{drummond03c4,he08learning,dong17class}.
Although the risk of overfitting decreases relative to oversampling, it has not been heavily explored due to the requirement to oust valuable information.

\textbf{Algorithm-based Approach.}
The algorithmic methods alter the learning or decision process in favor of minority concepts.
Most commonly, cost-sensitive techniques assign larger weights to the samples of minority concepts \cite{chen04using,tang08svms,zhou05training} in a way that
reduces the bias towards the majority.
This approach has the challenge of finding a suitable cost matrix because it has to be defined based on heuristics. Hence, intensive manual labor is a significant downside.

There do exist some hybrid methods of the data and algorithm approach by boosting techniques to ensemble the different partitions of the training distribution \cite{chawla03smoteboost}.
\cite{na19learning} and \cite{ye19learning} use what they call the episodic learning approach
that has the same underlying concept framed by deep learning techniques. They generate many episodes consisting of small unbalanced data and bootstrap the model from it.

With the rise of Deep learning, imbalanced learning has been receiving some limelight once more.
The methods broadly can be partitioned into three categories.
First, is transfer learning~\cite{bengio15sharing}. This technique purports to transfer the learned information from a larger pre-trained knowledge.
In the imbalanced problem, this would be from the majority concepts to the minority concepts and have been explored with modifications in \cite{lee16plankton,wang17nips,ouyang16cvpr,liu19large}.
Second is metric learning~\cite{huang16learning}. Which defines a metric space governed by objectives such as the triplet loss
where the goal is to group similar data and separate dissimilar data.
Third is few-shot~\cite{snell17prototype,vinyals17matching} and meta~\cite{finn17maml,ren18learning} learning. These techniques focus on generalizing from a small number of data, which perfectly aligns with the problem of imbalanced learning.

\begin{algorithm}[tb]
    \caption{Hierarchical class clustering}
    \label{algo:Hcc}

    \DontPrintSemicolon
    \begin{algorithmic}[1]
    {\footnotesize
    \REQUIRE $c_{I_{i}}$ $i \in N$(categories of image), $|c|$ (total number of categories), ngroups
    \ENSURE $G$ (set of clustered groups).
    
    \STATE \text{// Initialize set of groups}
    \STATE $G = \lbrace g_{1}, g_{2}, \cdots, g_{|c|}\rbrace$

    \WHILE{$len(G) > ngroups$}
        \STATE \text{// Initialize variables}
        \STATE $elem[g_{j}] = 0$ $\forall g_{j} \in G$ 
        \STATE $co[g_{j} \cup g_{k}] = 0. \forall g_{j}, g_{k} \in G$, $j \neq k$
        
        \STATE \text{// Calculate number of elements in each group}
        \FORALL{$i \in N$}
            \FORALL{$g_{j} \in G$}
                \IF{ $c_{I_{i}} \subseteq g_{j}$}
                    \STATE $elem[g_{j}]\mathrel{+}= 1$
                \ENDIF
            \ENDFOR
        \ENDFOR

        \STATE \text{// Calculate co-occurrence for each pair of groups}
        \FORALL{$i \in N$}
            \FORALL{$g_{j}, g_{k} \in G, j \neq k$}
                \IF{$c_{I_{i}} \not \subseteq g_{j}$, and $c_{I_{i}} \not \subseteq g_{k}$, and $c_{I_{i}} \subseteq (g_{i} \cup g_{j})$}
                    \STATE $co[g_{j} \cup g_{k}] \mathrel{+}= 1$
                \ENDIF
            \ENDFOR
        \ENDFOR

        \STATE \text{// Calculate merge score for each pair of groups}
        \FORALL{$g_{j}, g_{k} \in G, j \neq k$}
            \STATE $score_{merge}[g_{j} \cup g_{k}] = \ln{co[g_{j} \cup g_{k}]} - \beta(elem[g_{j}] + elem[g_{k}])^{2}$
        \ENDFOR

        \STATE \text{// Merge groups}
        \STATE $g_{a}, g_{b} = \argmax_{g_{j}, g_{k}} score_{merge}[g_{j} \cup g_{k}]$
        \STATE $G = G - g_{a} - g_{b} + (g_{a} \cup g_{b})$
        
    \ENDWHILE
    }
    \end{algorithmic}
\end{algorithm}

\subsection{Dataset Curation}
\label{sec:dataset_curation}
We enlist the curation details of the two sequential multi-labeled classification datasets, \textit{COCOseq} and \textit{NUS-WIDEseq}.
Also, we describe the process of sequential single-labeled classification dataset creation from MNIST~\cite{lecun98} and SVHN~\cite{netzer11}.

\subsubsection{Multi-Label Sequential Dataset Curation Details.}
\label{sec:multi_label_seq_dataset}
There are two objectives to satisfy when splitting a multi-label dataset into non-overlapping sub-tasks.
Recall, multi-labeled datasets inherently have intersecting concepts among the data points.
Hence, a naive splitting algorithm may lead to a dangerous amount of data loss.
This motivates our first objective to \textit{minimize} the data loss during the split. 
Additionally, in order for us to test diverse research environments,
the second objective is to optionally keep the size of the splits balanced.
To achieve the two objectives, we use the Hierarchical Clustering technique, which we tailor to maximize the amount of data while obtaining a desired balance among the tasks.
The algorithm merges classes in a bottom-up fashion until it reaches the designated number of groups.

The intuitive detail of algorithm~\ref{algo:Hcc} is as follows:
First, for each group (class or merged class set), count the number of images that \textit{only} have the labels which belong to its corresponding group (line 8--11).
Next, the algorithm explores by comparing the number of attainable images if two different groups are merged (line 13--16).
Finally, it selects the two groups, $g_j$ and $g_k$ to merge based on the following objective function,
\begin{align}
    score_{merge}[g_{j} \cup g_{k}] = \ln{co[g_{j} \cup g_{k}]} - \beta(elem[g_{j}] + elem[g_{k}])^{2} \label{eq:score_obj}
\end{align}
where $\beta$ is the inter-task balance control parameter.
Note that Eq.~\ref{eq:score_obj} assigns a higher score when merging $g_{j}$ and $g_{k}$ results in a larger number of images.
A lower score is assigned when the current sizes of $g_{j}$ and $g_{k}$ are already large.
The algorithm selects and merges the two groups with the higher score and iterates this process until only the desired number of groups remain.

During the curation  of \textit{COCOseq}, $\beta$ is set to 1 in order to promote the intra-task imbalance, but inter-task balance among the tasks. 
We imbue intra- \textit{and} inter-task imbalances in \textit{NUSWIDEseq}
by using a weakened $\beta$ of $0.0001$, with an additional constraint where the minimum number of classes for each of the tasks is set to 4.

\subsubsection{Single-Label Sequential Dataset Curation Details.}
\label{sec:single_label_seq_dataset}
To test the viability of PRS on benchmark single-label datasets, we first extend it to a long-tailed (Pareto distributed~\cite{reed01,newman05}) version.
Two benchmark datasets, MNIST~\cite{lecun98} and SVHN~\cite{netzer11} are explored. In accordance with \cite{wang17nips,liu19large}, we set the pareto distribution's power value, $\alpha=0.6$ to obtain \textit{Imbal-MNISTseq} and \textit{Imbal-SVHNseq}.
The \textit{Imbal-MNISTseq} consists of 11543 training and 10000 test set, while the \textit{Imbal-SVHNseq} consists of 14181 training and 26032 test set.
It is relatively straight forward to convert this into a sequential version, as each classes are perfectly disjoint in the single-labeled setting.

\begin{figure*}[tb]
    \begin{center}
    \includegraphics[width=\textwidth]{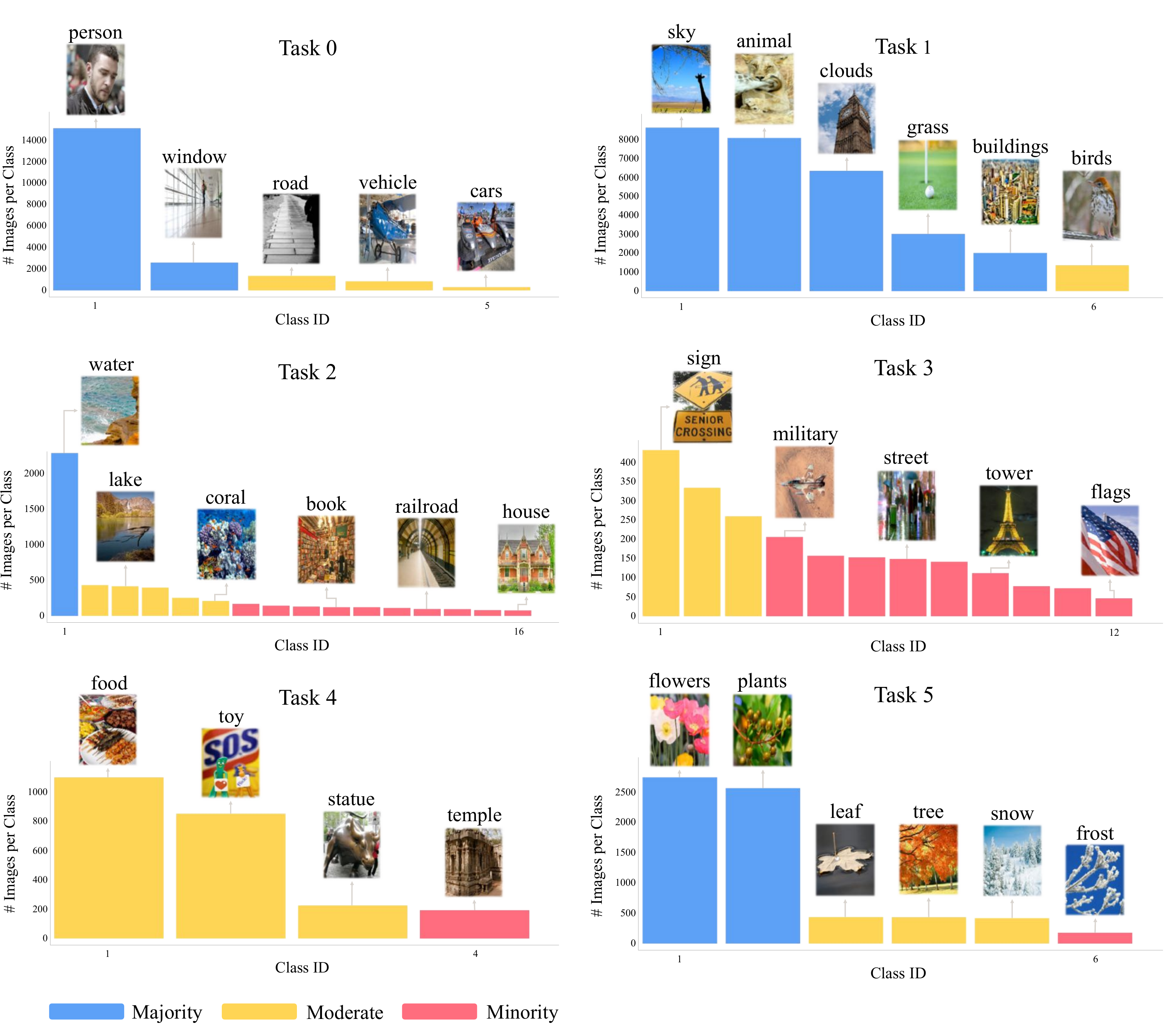}
    \caption{The \textit{NUSWIDEseq} dataset with 6 tasks. Note, it includes both intra- and inter-task imbalances. For instance, an inter-task imbalance is observed for \textit{Task 3} and \textit{Task 4}
        where, they both have much less number of data points compared to the other tasks}
    \label{fig:nuswide_sequence}
    \end{center}
\end{figure*}

\subsection{Experiments}
\label{sec:add_experiments}
In the supplemental experiment section, we enlist the details of how the base architectures and the continual baselines are trained. Also, the results of the RNN-Attention~\cite{wang17iccv},
and Attention consitency~\cite{guo19cvpr} experiments are reported along with the main table's OR, OP, CR, CP results and the task progression results of Table 1 from the main paper.
Additionally, we show the performance of PRS on single-labeled classification datasets \textit{Imbal-MNISTseq}, \textit{Imbal-SVHNseq}.

\subsubsection{Base Architectures.} We introduce the base architectures we explored during our experiments.\\

\textbf{Multi-labeled Classification.}
In addition to the plain ResNet101~\cite{he16cvpr}, we consider 2 other base architectures.
Due to the restrictions imposed by a task-free continual learning setting, the architectures cannot acquire any prior knowledge of the future inputs.
To that end, we choose Attention-Consistency~\cite{guo19cvpr} and RNN-Attention~\cite{wang17iccv}. Both approaches use either attention or recurrent techniques
to enhance the correlations of the labels. Table~\ref{tab:coco_nw_evalset} shows the results on all 3 base architectures on the multi-task setting~\cite{caruaca97}.
Notice, on the balanced test-set; all the base architectures perform much worse than its usual, imbalanced test-set. This phenomenon is observed whether if the model is trained for 1 or 10 epochs.
The results show when online trained (1 epoch), plain ResNet performs the best in the multi-task setting.
Furthermore, Table~\ref{tab:coco_ac} and Table~\ref{tab:coco_ra} also shows the performance on different architectures on \textit{COCOseq}.
We further verify that a plain ResNet performs best in a imbalanced online learning scenario, and importantly PRS always performs superior to other baselines no matter the architecture.

\textbf{Single-labeled Classification.}
For the \textit{Imbal-MNISTseq} experiments, we use a 2 layer fully connected network. The \textit{Imbal-SVHNseq} dataset uses a ResNet18~\cite{he16cvpr} as its base architecture.

\subsubsection{Baselines.}
Continual learning algorithms that satisfy our experimental setting, as well as the baselines that give a better relative understanding of the results, are listed.
We use a fixed online input batch size of $10$ in accordance with \cite{aljundi19gradient}, replay-batch size of 10, $\rho$ in [-0.2, 0.2], reservoir fixed to 2000, unless stated otherwise,
while performing only a single iteration per batch. Furthermore, all layers are finetuned unless stated otherwise and all the baselines were trained with the Adam
optimizer~\cite{kingma15} with $\beta_1=0.9$, $\beta_2=0.999$ and $\epsilon=1e-4$. We show 6 baselines, Multitask~\cite{caruaca97}, Finetune, EWC~\cite{kirkpatrick17ewc}, Conventional Reservoir Sampling~\cite{vitter85},
GSS-Greedy~\cite{aljundi19gradient}, ExStream~\cite{hayes19memory}.
\begin{enumerate}[label=\roman*.]
    \item \textbf{Multitask}~\cite{caruaca97}: i.i.d offline training for 1 epoch with a uniformly sampled minibatch.
    \item \textbf{Finetune}: Online training through the sequence of tasks.
    \item \textbf{EWC}~\cite{kirkpatrick17ewc}: Online training through the sequence of tasks with $\lambda$ optimized among 5 order of magnitudes and tuned to 15$k$ to show viable results.
        Importatly, in order to show sensible results on EWC, the ResNet had to be fixed up to the penultimate layer.%
    \item \textbf{Conventional Reservoir Sampling}~\cite{vitter85,riemer19iclr}: Online training with conventional reservoir sampling.
    \item \textbf{GSS-Greedy}~\cite{aljundi19gradient}: Online training with $10$ memories per task, memory strength of 10. %
    \item \textbf{ExStream}~\cite{hayes19memory}: Online training with the memory equally pre-partitioned using the prior distribution information of number of classes. 
        The pre-trained resnet is fixed up to the 2nd residual block to extract the features for their method prior to training.
        This was done because if the feature space changed due to finetuning, the performance degradation was larger.
    Also, since their method was designed for the single-label classification problem, we made a simplification assumption by assigning the label among the multiple as the fewest label in the current reservoir.
\end{enumerate}

\subsubsection{The Normalized Forgetting Measure.}
\label{subsec:forgettance}
Using~\cite{chaudhry18eccv}'s forgetting measure on an imbalanced setting may lead to imprecise measurements due to the relative differences in the
performance of the classes. Hence, we make a simple update to normalize the measure as shown,
\begin{align}
    F_k = \frac{1}{k-1} \sum_{j=1}^{k-1} f^k_j \\
    f^k_j = \max_{l\in{1,...,k-1}} \frac{m_{l,B_l,j} - m_{k, B_k, j}}{|m_{l,B_l,j}|}
\end{align}
where $f_j^k$ is the forgettance on task $j$ when the model is trained up to task $k$. $m$ is the metric we choose to measure forgetting with.

\begin{table*}[t!]
    \begin{minipage}{0.47\textwidth}
    \centering
    \scriptsize
    \setlength{\tabcolsep}{2pt}
    \caption{\textbf{\textit{Imbal-MNISTseq} results}.  The values represent the \textit{overall} performance after the completion of each task. 2 layer fully connected networks are used. All experiments are the mean of 5 random seeds}
    \label{tab:mnist_results}
    \begin{sc}
        \begin{tabular}{lccccc}
            \toprule
            & Task 1 & Task 2 & Task 3 & Task 4  & Task 5 \\
            \midrule
            MT      & - & - & - &  -   & 81.67 \\
            FT      & 21.47 & 19.66 & 10.24 & 11.1 & 9.82 \\
            CRS~\cite{vitter85}  & 21.43 & 39.22 & 33.7 & 43.93 & 52.57 \\
            GSS~\cite{aljundi19gradient}   & 21.44 & 39.59 & 44.57 & 48.96 & 62.54 \\
            PRS   & 21.46 & \textbf{40.04} & 31.3 & \textbf{57.84} & \textbf{68.45} \\
            \bottomrule
        \end{tabular}
    \end{sc}
\end{minipage}
\hspace{0.4cm}
\begin{minipage}{0.48\textwidth}
    \vskip 0.1in
    \centering
    \scriptsize
    \setlength{\tabcolsep}{2pt}
    \caption{\textbf{\textit{Imbal-SVHNseq} results}. The values represent the \textit{overall} performance after the completion of each task. The experiments use ResNet18~\cite{he16cvpr}.
    GSS~\cite{aljundi19gradient} experiment is the mean of 3 random seeds while all other experiments are the mean of 5 random seeds}
    \label{tab:svhn_results}
    \begin{sc}
        \begin{tabular}{lccccc}
            \toprule
            & Task 1 & Task 2 & Task 3 & Task 4  & Task 5 \\
            \midrule
            MT      & - & - & - &  -   & 70.68 \\
            FT  & 8.18 & 15.61 & 24.45 & 11.36 & 9.74 \\
            CRS~\cite{vitter85}  & 10.72 & 32.49 & 47.29 & 53.97 & 55.08 \\
            GSS~\cite{aljundi19gradient}  & 11.15 & 36.76 & 52.53 & 58.6 & 66.45 \\
            PRS  & 10.96 & 34.42 & \textbf{53.11} & \textbf{58.95} & \textbf{69.86} \\
            \bottomrule
        \end{tabular}
    \end{sc}
\end{minipage}
\end{table*}

\subsection{Results \& Analysis}
\label{sec:add_analysis}
In the supplemental analysis, we include the results on single-labeled classification, permuted \textit{NUSWIDEseq} schedules and memory sizes, as well as
analyze the gradient variance of the baselines in both \textit{Imbal-MNISTseq} and \textit{COCOseq}.

\subsubsection{Single-Labeled Classification Results.}
\label{sec:single_label_results}
Table~\ref{tab:mnist_results}, \ref{tab:svhn_results} show the results on \textit{Imbal-MNISTseq} and \textit{Imbal-SVHNseq} respectively.
Notice, even in the single-labeled classification setting, PRS proves very successful compared to all the other baselines~\cite{vitter85,riemer19iclr,aljundi19gradient}. 
Importantly, the PRS algorithm requires no alteration from the multi-labeled setting.
ExStream~\cite{hayes19memory} is omitted, due to the original paper and source code performing feature merging on a pre-trained feature space. 
It would be unfair to compare the other baselines if pre-trained data is only used for the ExStream baseline.

\subsubsection{Imbalanced test-set.}
To better understand PRS's capabilities, we further test its performance on the \textit{COCOseq}
imbalanced test-set obtained before the balancing
(Note that 50.5\% of MSCOCO's test-set belong to the majority classes).
Table~\ref{tab:nus_imb_evalset} shows PRS with $\rho=0.5$ performing better than
all the other baselines while remaining very competitive on the majority. \\

\subsubsection{Permuted \textit{NUSWIDEseq} Schedules.}
\label{sec:nw_schedule_permute}
In Table~\ref{tab:nw_schedule_permute_results}, we report 4 randomly permuted results of \textit{NUSWIDEseq}.
Notice, PRS performs the best while being relatively robust to the ordering of the tasks. The other baselines such as GSS~\cite{aljundi19gradient} and ExStream~\cite{hayes19memory}
show comparatively higher volatility in performance depending on the schedule.

\subsubsection{Different Memory Sizes on \textit{NUSWIDEseq}.}
\label{sec:nw_memory_sizes}
Table~\ref{tab:nw_memory_results} shows the results on \textit{NUSWIDEseq} on varying memory sizes (1000, 2000, 3000). On all three,
PRS remains as the superior model compared to the other baselines. Notably, there is a large jump in performance from 1000 to 2000 memory size,
but the gap quickly narrows when the memory grows from 2000 to 3000. This hints that in the online task-free setting, the memory size has a bound to its positive correlation with the performance.

\subsubsection{Task Progression Results.}
\label{sec:task_progression}
Table~\ref{tab:coco_results} and \ref{tab:nus-wide_results} shows the overall performance of the baselines and PRS as training progresses through each of the tasks.
The values indicate the performance right after the model finishes training on a task. We can observe that PRS has a good performance at any point in time of the training maintaining a
large margin between all the other baselines. Due to the maintenance property of PRS, it is able to support a well-rounded knowledge replays.
Hence, PRS should become more effective with increasing number of tasks. This trend is also evident in the single-label
classification on Table~\ref{tab:mnist_results}, \ref{tab:svhn_results}, where PRS starts to outperform after four tasks have been trained.

\begin{figure}[t!]
  \centering
  \begin{minipage}[t]{0.49\columnwidth}
    \includegraphics[width=\columnwidth]{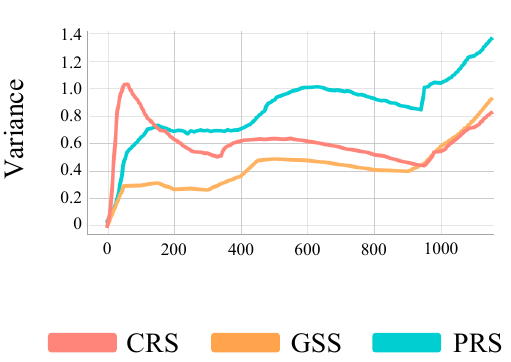}
    \caption{\textit{Imbal-MNISTseq} memory gradients variance comparison on CRS, GSS, and PRS. Each baseline is the mean of 5 random seeds}
    \label{fig:var_mnist}
  \end{minipage}
  \hfill
  \begin{minipage}[t]{0.49\columnwidth}
    \includegraphics[width=\columnwidth]{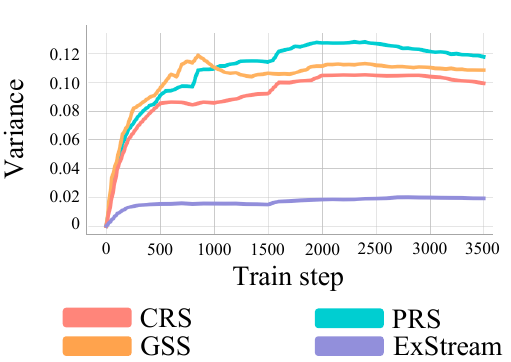}
    \caption{\textit{COCOseq} memory gradients variance comparison on CRS, GSS, ExStream and PRS. GSS is the mean of 3 random seeds while other baselines are the mean of 5 random seeds.
    Note that ExStream has a different training method requiring pre-featurized inputs and fixing some parameters which complicates the comparison}
    \label{fig:var_coco}
  \end{minipage}
\end{figure}

\subsubsection{Gradient Variance.}
\label{sec:grad_var}
An interesting phenomenon is observed in the training dynamics, specifically in the memory's gradient variances. 
As done in \cite{greensmith04jmlr,papini18a,xu18aistats}, the variance value is computed by the $Trace$ of the gradient covariance matrix,
\begin{align}
Tr(\mathbb{E}[(g-\mathbb{E}[g])(g-\mathbb{E}[g])^\intercal])
\end{align}
where $g$ is our gradient vector. In Figures~\ref{fig:var_mnist} and \ref{fig:var_coco}, we compare the gradient variance of the models
trained on \textit{Imbal-MNISTseq} and \textit{COCOseq}, respectively. A clear pattern we discover is that PRS maintains a
relatively higher gradient variance compared to the other baselines. A possible explanation of this phenomenon is that PRS’s
maintenance algorithm is able to guarantee the diversity of the classes in the memory throughout training.
Which in other words means PRS is continuously working to optimize towards satisfying a higher number of objectives.
This naturally leads to higher gradient variances which could in fact be an unintended side-effect of preventing the forgetting of the past experiences.

\begin{table*}
    \centering
    \scriptsize
    \setlength{\tabcolsep}{2pt}
    \caption{\textbf{\textit{COCOseq} \& \textit{NUS-WIDEseq}, comparison of imbalanced evaluation set performance and balanced set performance} on ResNet~\cite{he16cvpr}, Recurrent Attention (RA)~\cite{wang17iccv} 
        and \cite{guo19cvpr}'s Attention Consistency (AC).
    Note, due to the system's memory limitations, AC had to freeze the first residual block. For a fair comparison, the same is done on the plain ResNets in this table. Additionally, to obtain sensible results 
    we separately pretrain the RA on ImageNet.
    All results are the mean of 5 repeats.}
    \label{tab:coco_nw_evalset}
    \begin{sc}
        \begin{tabular}{lccc|ccc}
            \toprule
            &    \multicolumn{3}{c|}{\textit{NUS-WIDEseq} overall} & \multicolumn{3}{c}{\textit{COCOseq} overall}\\
                                      & C-F1\hspace{0.2cm} &  O-F1\hspace{0.2cm} &  MAP\hspace{0.2cm} & C-F1\hspace{0.2cm} &  O-F1\hspace{0.2cm} &  MAP\hspace{0.2cm} \\
            \midrule
            Res-MT-1epc(imbal)       & 33.5 &  69.8 & 34.4 & 54.7 & 67.7  & 58.2 \\
            Res-MT-10epc(imbal)      & 39.9 &  69.2 & 37.1 & 61.7 & 71.8  & 63.7 \\
            \specialrule{0.1pt}{1pt}{1pt}
            Res-MT-1epc(bal)         & 26.0 &  24.8 & 28.4 & 55.3 & 54.7  & 56.8 \\
            Res-MT-10epc(bal)        & 35.9 &  31.3 & 32.7 & 60.6 & 60.8  & 62.6 \\
            \specialrule{0.1pt}{1pt}{1pt}

            RA-MT-1epc(imbal)       & 21.5 &  70.9 & 30.1 & 39.4 & 65.4  & 55.8 \\
            RA-MT-10epc(imbal)      & 39.3 &  70.6 & 37.5 & 70.2 & 79.9  & 73.3 \\
            \specialrule{0.1pt}{1pt}{1pt}
            RA-MT-1epc(bal)         & 13.4 &  19.5 & 22.6 & 39.2 & 47.2  & 52.2 \\
            RA-MT-10epc(bal)        & 34.5 &  29.8 & 32.4 & 68.7 & 69.3  & 70.9 \\
            \specialrule{0.1pt}{1pt}{1pt}

            AC-MT-1epc(imbal)        &  26.2 & 71.7 & 31.3 & 46.0 & 64.2  & 51.2 \\
            AC-MT-10epc(imbal)       & 41.2 &  74.0 & 40.7 & 65.8 & 77.7  & 68.9 \\
            \specialrule{0.1pt}{1pt}{1pt}
            AC-MT-1epc(bal)          &  16.5 & 20.0 & 25.7 & 42.5 & 47.0  & 47.4 \\
            AC-MT-10epc(bal)         & 33.6 &  31.1 & 33.7 & 61.7 & 63.0  & 65.3 \\
            \bottomrule
        \end{tabular}
    \end{sc}
    \vskip -0.2in
\end{table*}

\begin{table*}[t!]
    \centering
    \scriptsize
    \setlength{\tabcolsep}{2pt}
    \caption{\textbf{Imbalanced test-set comparison} on \textit{COCOseq}. Please refer to Table~\ref{tab:coco_results} for the nomenclatures and experimental detail. $\rho$ is set to $0.5$.
    }
    \label{tab:nus_imb_evalset}
    \begin{sc}
        \begin{tabular}{lccc|ccc|ccc||ccc}
            \toprule
            &\multicolumn{3}{c|}{majority} & \multicolumn{3}{c|}{moderate}& \multicolumn{3}{c||}{minority}& \multicolumn{3}{c}{\textbf{Overall}}\\
            \textbf{\textit{COCOseq}}      & C-F1 & O-F1 & mAP  & C-F1 & O-F1 & mAP  & C-F1 & O-F1 &  mAP & C-F1 &  O-F1 &  mAP \\
            \midrule

            CRS~\cite{vitter85} & \textcolor{red}{72.3} & \textcolor{red}{69.6} & \textcolor{red}{78.7} & \textcolor{red}{52.0} & \textcolor{blue}{52.3} & \textcolor{blue}{57.0} & 14.2 & 16.4 & 20.7 & \textcolor{blue}{50.8} & \textcolor{blue}{60.1} & \textcolor{blue}{55.3} \\
            GSS~\cite{aljundi19gradient} & 64.7 & 63.8 & 70.2 & 47.3 & 49.0 & 49.8 & 9.1 & 9.5 & 10.6 & 44.5 & 54.8 & 47.2 \\
            ExStream~\cite{hayes19memory}& 63.2 & 56.6 & 72.9 & 49.5 & 46.8 & 55.5 & \textcolor{blue}{22.2} & \textcolor{blue}{23.0} & \textcolor{blue}{27.0} & 48.3 & 50.9 & 54.3 \\
            \specialrule{0.4pt}{1pt}{1pt}
            PRS(ours)& \textcolor{blue}{71.9} & \textcolor{blue}{69.2} & \textcolor{blue}{78.6} & \textcolor{red}{52.0} & \textcolor{red}{52.6} & \textcolor{red}{58.8} & \textcolor{red}{23.0} & \textcolor{red}{24.4} & \textcolor{red}{27.3} & \textcolor{red}{52.4} & \textcolor{red}{60.4} & \textcolor{red}{57.6} \\

            \bottomrule
        \end{tabular}
    \end{sc}
    \vskip -0.2in
\end{table*}

\begin{table*}
    \centering
    \scriptsize
    \setlength{\tabcolsep}{2pt}
    \caption{\textbf{CP, CR, OP, OR \textit{COCOseq} results}. Additional evaluation metrics for multi-labeled learning are reported. Similar to ~\cite{liu19large},
    we report the majority, moderate and minority of the distribution to accurately assess the long-tail performances. MT refers to ``Multitask", FT refers to ``finetune",
    CRS~\cite{vitter85}, GSS~\cite{aljundi19gradient}, and our PRS method. FORGET refers to the Normalized forgetting measure~\cite{chaudhry18eccv}. Due to the large time complexity, GSS~\cite{aljundi19gradient} result is the mean of 3 random seeds
    while the rest are the mean of 5 random seeds. In \textcolor{red}{red} and \textcolor{blue}{blue} are the \textbf{$1^{st}$} and \textbf{$2^{nd}$} best performances, respectively.}
    \label{tab:coco_cpcropor_results}
    \begin{sc}
        \begin{tabular}{lcccc|cccc|cccc}
            \toprule
            &\multicolumn{4}{c|}{majority}    & \multicolumn{4}{c|}{moderate} & \multicolumn{4}{c}{minority} \\
            & C-P\hspace{0.2cm} & C-R\hspace{0.2cm} & O-P\hspace{0.2cm} & O-R\hspace{0.2cm}  & C-P\hspace{0.2cm} & C-R\hspace{0.2cm} & O-P\hspace{0.2cm} & O-R\hspace{0.2cm} & C-P\hspace{0.2cm} & C-R\hspace{0.2cm} & O-P\hspace{0.2cm} & O-R\hspace{0.2cm} \\
            \midrule
            MT       & 72.2 & 73.7 & 68.5 & 73.7 & 55.4 & 51.3 & 51.7 & 51.3 & 28.3 & 8.6 & 42.2 & 8.6\\
            \midrule
            FT       & 13.3 & 30.5 & 25.7 & 30.5 & 4.4 & 14.3 & 20.0 & 14.3 & 0.0 & 0.0 & 0.0 & 0.0 \\
            EWC      & 55.5 & \textcolor{blue}{65.4} & 45.1 & \textcolor{blue}{65.4} & 41.8 & 33.7 & 43.9 & 33.7 & 26.5 & 4.4 & \textcolor{blue}{74.6} & 4.4 \\
            CRS      & \textcolor{blue}{68.2} & \textcolor{red}{65.8} & \textcolor{red}{59.5} & \textcolor{red}{65.8} & 61.4 & \textcolor{blue}{39.1} & \textcolor{blue}{53.5} & \textcolor{blue}{39.1} & 38.9 & 9.1 & 59.6 & 9.1 \\
            GSS      & 64.6 & 56.1 & 57.7 & 56.1 & 56.4 & 38.1 & 51.2 & 38.1 & 30.0 & 6.3 & 38.5 & 6.3 \\
            ExStream & 66.9 & 52.6 & 51.4 & 52.6 & \textcolor{blue}{63.4} & 38.0 & 52.1 & 38.0 & \textcolor{red}{72.8} & \textcolor{blue}{16.2} & \textcolor{red}{76.0} & \textcolor{blue}{16.2} \\
            \specialrule{0.4pt}{1pt}{1pt}
            PRS(ours) & \textcolor{red}{70.8} & 60.8 & \textcolor{blue}{57.9} & 60.8 &
                        \textcolor{red}{67.8} & \textcolor{red}{42.8} & \textcolor{red}{59.3} & \textcolor{red}{42.8} &
                        \textcolor{blue}{65.9} & \textcolor{red}{23.4} & 66.7 & \textcolor{red}{23.4} \\

            \bottomrule
        \end{tabular}
    \end{sc}
\end{table*}

\begin{table*}
    \centering
    \scriptsize
    \setlength{\tabcolsep}{2pt}
    \caption{\textbf{CP, CR, OP, OR \textit{NUSWIDEseq} results}. Please refer to table~\ref{tab:coco_cpcropor_results} for the nomenclatures and details of the experiments}

    \label{tab:nw_cpcropor_results}
    \begin{sc}
        \begin{tabular}{lcccc|cccc|cccc}
            \toprule
            &\multicolumn{4}{c|}{majority}    & \multicolumn{4}{c|}{moderate} & \multicolumn{4}{c}{minority} \\
            & C-P\hspace{0.2cm} & C-R\hspace{0.2cm} & O-P\hspace{0.2cm} & O-R\hspace{0.2cm}  & C-P\hspace{0.2cm} & C-R\hspace{0.2cm} & O-P\hspace{0.2cm} & O-R\hspace{0.2cm} & C-P\hspace{0.2cm} & C-R\hspace{0.2cm} & O-P\hspace{0.2cm} & O-R\hspace{0.2cm} \\
            \midrule
            MT      & 25.4 & 50.0 & 22.3 &  50.0  &  38.7 & 23.8 &  36.7 &  23.8 & 21.1 & 6.6 & 63.6  & 6.6 \\
            \midrule
            FT      & 0.3 & 7.6 & 3.4 & 7.6 & 4.4 & 1.5 & 22.6 & 1.5 & 6.7 & 4.3 & 28.0 & 4.3 \\

            EWC  & 10.1 & \textcolor{red}{35.6} & 5.8 & \textcolor{red}{35.6} & 21.0 & \textcolor{red}{13.3} & 11.9 & \textcolor{red}{13.3} & 20.4 & 9.1 & 31.7 & 9.1 \\

            CRS  & \textcolor{blue}{30.1} & \textcolor{blue}{27.0} & \textcolor{blue}{13.2} & \textcolor{blue}{27.0} & 29.3 & 9.4 & 37.2 & 9.4 & 27.3 & 8.4 & 44.0 & 8.4 \\

            GSS  & 28.4 & 22.0 & 9.8 & 22.0 & 26.5 & 10.3 & 31.2 & 10.3 & 33.7 & 10.4 & \textcolor{red}{56.8} & 10.4 \\

            ExStream & 24.3 & 11.7 & 7.8 & 11.7 & \textcolor{blue}{33.6} & 7.7 & \textcolor{blue}{38.3} & 7.7 & \textcolor{blue}{41.1} & \textcolor{blue}{17.7} & 38.4 & \textcolor{blue}{17.7} \\
            \specialrule{0.4pt}{1pt}{1pt}
            PRS & \textcolor{red}{34.8} & 21.8 & \textcolor{red}{15.2} & 21.8 & \textcolor{red}{38.2} & \textcolor{blue}{13.0} & \textcolor{red}{38.6} & \textcolor{blue}{13.0} & \textcolor{red}{57.7} & \textcolor{red}{18.1} & \textcolor{blue}{52.2} & \textcolor{red}{18.1} \\

            \bottomrule
        \end{tabular}
    \end{sc}
\end{table*}

\begin{table*}
    \centering
    \scriptsize
    \setlength{\tabcolsep}{2pt}
    \caption{\textbf{Per task \textit{COCOsequence} results with AC~\cite{guo19cvpr} architecture}. Conventional evaluation metrics for multi-labeled
    learning are reported after the whole data stream is trained. Similar to \cite{liu19large},
    the majority, moderate and minority are reported to accurately assess the long-tail performances.
    \textit{full} refers to the total test-set being used to evaluate the model after training all the tasks. 
    Each $T=[0,1,2,3]$ refers to the current task test-set performance right after training on it.
    Due to the large time complexity, GSS-Greedy~\cite{aljundi19gradient}'s result is the mean of 3 experiments
    while the rest are the mean of 5. In \textcolor{red}{red} and \textcolor{blue}{blue} are the \textbf{$1^{st}$} and \textbf{$2^{nd}$} best performances, respectively.
    FORGET refers to the normalized forgetting measure~\cite{chaudhry18eccv}}
    \label{tab:coco_ac}
    \begin{sc}
        \begin{tabular}{lccc|ccc|ccc||ccc}
            \toprule
            &\multicolumn{3}{c|}{majority} & \multicolumn{3}{c|}{moderate}& \multicolumn{3}{c||}{minority}& \multicolumn{3}{c}{overall}\\
                     & C-F1 & O-F1 & MAP  & C-F1 & O-F1 & MAP  & C-F1 & O-F1 &  MAP & C-F1 &  O-F1 &  MAP \\
            \midrule
            MT      & 68.8 & 65.3 & 70.2 &  44.0  &  45.9 & 46.4 &  0.4 &   0.4 & 20.6 & 42.5 & 47.0  & 47.4 \\

            \midrule
            FT(full)& 14.2 & 24.1 & 34.2 & 5.1 & 13.2 & 19.7 & 0.0 & 0.0 & 8.4 & 6.4 & 15.5 & 21.1 \\
            \;forget & 100.0 & 100.0 & 38.6 & 100.0 & 100.0 & 32.7 & 100.0 & 100.0 & 26.0 & 100.0 & 100.0 & 32.9 \\

            \midrule
            CRS(full)& \textcolor{red}{65.0} & \textcolor{red}{59.4} & \textcolor{red}{62.6} & 44.2 & \textcolor{blue}{42.0} & 44.8 & 6.3 & 6.9 & 24.2 & 43.8 & \textcolor{blue}{43.6} & 45.2 \\
            \;forget & 7.2  & 6.4  & 7.3  & 25.4 & 21.4 & 4.1  & 61.9 & 63.1 & 8.6 & 22.0 & 19.0 & 5.5 \\

            \midrule
            GSS(full)& 63.9 & \textcolor{blue}{59.0} & \textcolor{blue}{62.3} & 42.0 & 41.8 & \textcolor{blue}{45.6} & 5.5 & 6.6 & 22.8 & 41.2 & 43.2 & \textcolor{blue}{45.3} \\
            \;forget & 9.0 & 7.1 & 5.2 & 34.1 & 30.8 & 0.7 & 71.1 & 72.3 & 6.5 & 26.9 & 22.6 & 2.7 \\

            \midrule
            Exstream(full)& 56.0 & 46.4 & 55.7 & \textcolor{blue}{45.5} & 39.4 & 44.2 & \textcolor{blue}{29.1} & \textcolor{blue}{29.0} & \textcolor{blue}{30.2} & \textcolor{blue}{45.9} & 40.0 & 44.3 \\
            \;forget & 34.2 & 32.5 & 22.6 & 30.9 & 29.1 & 18.0 & 37.4 & 36.1 & 16.6 & 33.2 & 31.3 & 18.5 \\

            \midrule
            PRS(full)& \textcolor{blue}{64.2} & \textcolor{blue}{59.0} & 61.9 & \textcolor{red}{49.9} & \textcolor{red}{48.0} & \textcolor{red}{50.6} & \textcolor{red}{35.9} & \textcolor{red}{35.4} & \textcolor{red}{37.3} & \textcolor{red}{51.4} & \textcolor{red}{49.2} & \textcolor{red}{50.8} \\
            \;forget & 15.8 & 14.1 & 6.0 & 18.9 & 18.9 & 2.6 & 6.6 & 6.4 & 5.2 & 17.3 & 16.7 & 3.6 \\
            \bottomrule
        \end{tabular}
    \end{sc}
\end{table*}

\begin{table*}
    \centering
    \scriptsize
    \setlength{\tabcolsep}{2pt}
    \caption{\textbf{\textit{COCOsequence} results with RA~\cite{wang17iccv} architecture}. Note that GSS is omitted due to system memory limits.
    Please refer to table~\ref{tab:coco_ac} for the nomenclatures and details of the experiments}
    \label{tab:coco_ra}
    \begin{sc}
        \begin{tabular}{lccc|ccc|ccc||ccc}
            \toprule
            &\multicolumn{3}{c|}{majority} & \multicolumn{3}{c|}{moderate}& \multicolumn{3}{c||}{minority}& \multicolumn{3}{c}{overall}\\
                     & C-F1 & O-F1 & MAP  & C-F1 & O-F1 & MAP  & C-F1 & O-F1 &  MAP & C-F1 &  O-F1 &  MAP \\
            \midrule
            MT      & 71.1 & 70.1 & 76.2 &  25.9  &  32.2 & 45.2 & 0.0 &  0.0 & 15.7 & 33.0 & 41.6  & 47.3 \\

            \midrule
            FT(full)& 19.3 & 29.4 & 32.1 & 8.1 & 17.9 & 17.0 & 0.0 & 0.0 & 4.2 & 9.4 & 19.4 & 18.3 \\
            \;forget & 100.0 & 100.0 & 52.5 & 100.0 & 100.0 & 53.4 & 100.0 & 100.0 & 38.5 & 100.0 & 100.0 & 50.7 \\

            \midrule
            CRS(full)& \textcolor{red}{69.1} & \textcolor{blue}{64.4} & \textcolor{red}{69.9} & 47.4 & 46.7 & 49.7 & 6.9 & 7.6 & 21.8 & 46.4 & 47.6 & 49.3 \\
            \;forget & 11.6 & 10.8 & 7.3 & 16.2 & 19.7 & 8.8 & -24.9 & -25.6 & 10.7 & 9.9 & 16.1 & 7.7 \\

            \midrule
            ExStream(full)& 62.7 & 56.1 & 64.2 & \textcolor{blue}{50.8} & \textcolor{blue}{48.8} & \textcolor{blue}{52.7} & \textcolor{blue}{28.4} & \textcolor{blue}{29.2} & \textcolor{red}{35.9} & \textcolor{blue}{50.6} & \textcolor{blue}{48.2} & \textcolor{blue}{52.3} \\
            \;forget & 35.7 & 35.0 & 18.1 & 31.0 & 31.0 & 14.3 & 18.8 & 21.1 & 14.4 & 31.4 & 31.7 & 15.1 \\

            \midrule
            PRS(full)& \textcolor{blue}{68.5} & \textcolor{red}{64.5} & \textcolor{blue}{69.4} & \textcolor{red}{51.8} & \textcolor{red}{51.3} & \textcolor{red}{55.1} & \textcolor{red}{29.1} & \textcolor{red}{30.6} & \textcolor{blue}{34.5} & \textcolor{red}{52.5} & \textcolor{red}{52.1} & \textcolor{red}{54.6} \\
            \;forget & 19.4 & 19.1 & 7.5 & 20.9 & 21.5 & 4.8 & -16.2 & -12.8 & 2.1 & 15.5 & 17.4 & 4.1 \\
            \bottomrule
        \end{tabular}
    \end{sc}
\end{table*}

\begin{table*}
    \centering
    \scriptsize
    \setlength{\tabcolsep}{2pt}
    \caption{\textbf{\textit{NUS-WIDEseq} performance on different reservoir sizes}. Please refer to table~\ref{tab:coco_ac} for the nomenclatures and details of the experiments.}
    \label{tab:nw_memory_results}
    \begin{sc}
        \begin{tabular}{lccc|ccc|ccc||ccc}
            \toprule
            &\multicolumn{3}{c|}{majority} & \multicolumn{3}{c|}{moderate}& \multicolumn{3}{c||}{minority}& \multicolumn{3}{c}{overall}\\
                     & C-F1 & O-F1 & MAP  & C-F1 & O-F1 & MAP  & C-F1 & O-F1 &  MAP & C-F1 &  O-F1 &  MAP \\
            \toprule
                     \multicolumn{13}{c}{reservoir: $1000$}\\
            \midrule
            CRS(full)& \textcolor{red}{28.1} & \textcolor{red}{15.1} & \textcolor{red}{20.0} & 11.2 & 11.5 & 17.1 & 11.3 & 12.6 & 19.1 & 15.3 & 13.3 & 18.6 \\
            \;forget & 44.5 & 41.3 & 16.1 & 74.4 & 72.6 & 18.7 & 98.5 & 98.5 & 30.5 & 68.3 & 65.0 & 20.2 \\

            GSS(full)& \textcolor{blue}{24.8} & \textcolor{blue}{14.9} & \textcolor{blue}{18.6} & \textcolor{blue}{13.9} & \textcolor{blue}{14.6} & 17.2 & 10.8 & 11.5 & 20.8 & 16.2 & \textcolor{blue}{13.9} & 19.0 \\
            \;forget & 39.1 & 39.0 & 16.7 & 65.6 & 63.6 & 13.6 & 90.7 & 90.4 & 17.0 & 58.9 & 55.4 & 17.4 \\

            ExStream(full)& 16.5 & 8.1 & 15.6 & 11.8 & 11.8 & \textcolor{blue}{17.3} & \textcolor{blue}{20.9} & \textcolor{blue}{20.9} & \textcolor{blue}{24.1} & \textcolor{blue}{16.9} & 13.8 & \textcolor{blue}{19.9} \\
            \;forget & 80.5 & 77.3 & 22.7 & 90.7 & 90.6 & 25.8 & 89.2 & 89.0 & 23.2 & 87.6 & 86.1 & 24.2 \\

            PRS(full)& 19.3 & 11.0 & 18.2 & \textcolor{red}{15.1} & \textcolor{red}{15.4} & \textcolor{red}{18.9} & \textcolor{red}{24.1} & \textcolor{red}{22.9} & \textcolor{red}{26.7} & \textcolor{red}{20.1} & \textcolor{red}{16.8} & \textcolor{red}{22.1} \\
            \;forget & 70.4 & 67.3 & 18.2 & 74.4 & 74.1 & 15.8 & 68.8 & 68.4 & 13.7 & 71.0 & 69.4 & 15.4 \\
            \midrule
                     \multicolumn{13}{c}{reservoir: $2000$}\\
            \midrule
            CRS(full)& \textcolor{red}{28.1} & \textcolor{blue}{17.6} & \textcolor{red}{21.4} & 14.1 & 15.0 & \textcolor{blue}{18.5} & 12.8 & 13.9 & 21.3 & 17.5 & 15.7 & 20.3 \\
            \;forget & 36.0 & 32.5 & 13.8 & 62.3 & 60.7 & 15.7 & 95.5 & 95.4 & 21.6 & 60.3 & 57.3 & 17.1 \\

            GSS(full)& 24.6 & 13.5 & 19.0 & \textcolor{blue}{14.8} & \textcolor{blue}{15.5} & 17.9 & 15.9 & 17.6 & 24.5 & 17.9 & 15.3 & 20.9 \\
            \;forget & 46.8 & 43.9 & 16.6 & 59.6 & 58.3 & 11.7 & 82.8 & 82.0 & 18.6 & 54.8 & 49.8 & 13.0 \\

            ExStream(full)& 15.6 & 9.2 & 15.3 & 12.4 & 12.8 & 17.6 & \textcolor{blue}{24.6} & \textcolor{blue}{24.1} & \textcolor{blue}{26.7} & \textcolor{blue}{18.7} & \textcolor{blue}{16.0} & \textcolor{blue}{21.0} \\
            \;forget & 80.7 & 77.6 & 24.0 & 81.0 & 80.6 & 23.3 & 77.2 & 76.7 & 21.8 & 81.0 & 79.3 & 23.4 \\

            PRS(full)& \textcolor{blue}{26.7} & \textcolor{red}{17.9} & \textcolor{blue}{21.2} & \textcolor{red}{19.2} & \textcolor{red}{19.3} & \textcolor{red}{21.5} & \textcolor{red}{27.5} & \textcolor{red}{26.8} & \textcolor{red}{31.0} & \textcolor{red}{24.8} & \textcolor{red}{21.7} & \textcolor{red}{25.5} \\
            \;forget & 45.8 & 43.0 & 15.7 & 59.0 & 58.4 & 13.4 & 60.6 & 60.3 & 15.5 & 55.3 & 53.5 & 13.9 \\

            \midrule
                     \multicolumn{13}{c}{reservoir: $3000$}\\
            \midrule
            CRS(full)& \textcolor{red}{29.6} & \textcolor{red}{20.2} & \textcolor{blue}{22.1} & \textcolor{blue}{15.1} & \textcolor{blue}{15.9} & \textcolor{blue}{18.8} & 12.9 & 14.0 & 23.0 & 18.5 & \textcolor{blue}{16.9} & 21.3 \\
            \;forget & 29.7 & 27.0 & 16.1 & 52.7 & 50.9 & 15.0 & 92.3 & 91.9 & 18.8 & 52.9 & 49.1 & 16.9 \\

            GSS(full)& 26.2 & 17.2 & 19.2 & 11.8 & 12.7 & 16.6 & 14.9 & 17.0 & 24.1 & 17.2 & 15.8 & 20.3 \\
            \;forget & 34.9 & 30.4 & 15.0 & 66.8 & 65.5 & 16.2 & 81.6 & 80.6 & 6.2 & 54.2 & 49.2 & 13.6 \\

            ExStream(full)& 17.9 & 9.9 & 16.8 & 12.9 & 13.3 & 18.4 & \textcolor{blue}{26.3} & \textcolor{blue}{25.1} & \textcolor{blue}{28.0} & \textcolor{blue}{20.0} & 16.6 & \textcolor{blue}{22.2} \\
            \;forget & 78.1 & 74.3 & 25.3 & 77.0 & 76.7 & 20.5 & 76.6 & 76.4 & 25.4 & 78.6 & 77.2 & 22.6 \\

            PRS(full)& \textcolor{blue}{26.7} & \textcolor{blue}{19.7} & \textcolor{red}{22.3} & \textcolor{red}{20.4} & \textcolor{red}{20.3} & \textcolor{red}{22.4} & \textcolor{red}{27.7} & \textcolor{red}{27.9} & \textcolor{red}{31.1} & \textcolor{red}{25.3} & \textcolor{red}{23.0} & \textcolor{red}{26.1} \\
            \;forget & 43.6 & 40.3 & 12.3 & 48.4 & 47.2 & 14.5 & 41.6 & 41.2 & 10.0 & 48.1 & 46.0 & 12.2 \\

            \bottomrule
        \end{tabular}
    \end{sc}
\end{table*}

\begin{table*}
    \centering
    \scriptsize
    \setlength{\tabcolsep}{2pt}
    \caption{\textbf{\textit{NUS-WIDEseq} performance on permuted schedules.} Please refer to table~\ref{tab:coco_ac} for the nomenclatures and details of the experiments.}
    \label{tab:nw_schedule_permute_results}
    \begin{sc}
        \begin{tabular}{lccc|ccc|ccc||ccc}
            \toprule
            &\multicolumn{3}{c|}{majority} & \multicolumn{3}{c|}{moderate}& \multicolumn{3}{c||}{minority}& \multicolumn{3}{c}{overall}\\
                     & C-F1 & O-F1 & MAP  & C-F1 & O-F1 & MAP  & C-F1 & O-F1 &  MAP & C-F1 &  O-F1 &  MAP \\
            \toprule
                     \multicolumn{13}{c}{Schedule: 0, 1, 2, 3 , 4, 5}\\
            \midrule
            CRS(full)& \textcolor{red}{27.4} & \textcolor{red}{23.1} & \textcolor{red}{21.6} & \textcolor{blue}{15.1} & \textcolor{blue}{15.9} & \textcolor{blue}{16.0} & 3.8 & 4.3 & 15.6 & \textcolor{blue}{14.6} & \textcolor{blue}{15.2} & 16.9 \\
            \;forget & 26.0 & 24.8 & 18.0 & 57.8 & 55.3 & 20.1 & 89.2 & 89.5 & 17.9 & 58.1 & 54.9 & 17.3 \\

            GSS(full)& 25.0 & \textcolor{blue}{22.0} & 19.6 & 9.4 & 9.8 & 14.9 & 1.0 & 1.1 & 13.8 & 10.3 & 11.3 & 15.4 \\
            \;forget & 43.1 & 41.9 & 14.9 & 59.9 & 60.5 & 12.8 & 94.1 & 94.6 & 15.9 & 63.0 & 62.9 & 14.3 \\

            ExStream(full)& 14.2 & 12.7 & 14.2 & 12.4 & 10.3 & 15.2 & \textcolor{blue}{10.1} & \textcolor{blue}{10.1} & \textcolor{blue}{21.7} & 13.1 & 11.0 & \textcolor{blue}{17.8} \\
            \;forget & 95.7 & 95.5 & 27.5 & 85.1 & 84.9 & 22.0 & 74.5 & 74.4 & 13.6 & 86.8 & 86.0 & 20.2 \\

            PRS(full)& \textcolor{blue}{25.1} & 20.9 & \textcolor{blue}{21.5} & \textcolor{red}{19.8} & \textcolor{red}{18.8} & \textcolor{red}{20.4} & \textcolor{red}{24.8} & \textcolor{red}{24.7} & \textcolor{red}{29.5} & \textcolor{red}{23.4} & \textcolor{red}{21.6} & \textcolor{red}{24.5} \\
            \;forget & 72.9 & 71.4 & 15.6 & 61.6 & 61.3 & 13.0 & 24.3 & 24.4 & 2.3 & 51.4 & 49.1 & 7.8 \\
            \midrule
                     \multicolumn{13}{c}{Schedule: 1, 2, 0, 5, 3, 4}\\
            \midrule
            CRS(full)& \textcolor{red}{25.9} & \textcolor{red}{25.1} & \textcolor{blue}{21.1} & \textcolor{blue}{16.9} & \textcolor{blue}{14.4} & \textcolor{blue}{17.5} & 3.6 & 3.6 & 16.5 & \textcolor{blue}{13.9} & \textcolor{blue}{13.8} & \textcolor{blue}{17.8} \\
            \;forget & 29.2 & 25.8 & 16.7 & 65.0 & 65.7 & 18.4 & 92.0 & 92.1 & 16.3 & 60.6 & 58.3 & 19.4 \\

            GSS(full)& \textcolor{blue}{23.4} & 20.9 & 18.4 & 10.3 & 9.6 & 14.3 & 2.7 & 2.8 & 13.2 & 10.5 & 10.4 & 14.6 \\
            \;forget & 29.4 & 27.8 & 8.6 & 80.3 & 79.7 & 17.7 & 93.2 & 94.0 & 20.4 & 68.9 & 65.7 & 19.0 \\

            ExStream(full)& 7.9 & 8.8 & 14.2 & 16.5 & 11.3 & 15.9 & \textcolor{blue}{8.6} & \textcolor{blue}{7.1} & \textcolor{blue}{19.5} & 11.7 & 9.3 & 17.1 \\
            \;forget & 82.1 & 79.8 & 23.2 & 87.2 & 86.7 & 26.3 & 64.1 & 63.8 & 18.9 & 82.9 & 81.5 & 25.9 \\

            PRS(full)& 20.9 & \textcolor{blue}{21.8} & \textcolor{red}{21.5} & \textcolor{red}{21.7} & \textcolor{red}{19.0} & \textcolor{red}{21.2} & \textcolor{red}{23.9} & \textcolor{red}{22.7} & \textcolor{red}{29.7} & \textcolor{red}{22.9} & \textcolor{red}{21.0} & \textcolor{red}{24.9} \\
            \;forget & 55.2 & 52.4 & 10.4 & 58.1 & 57.5 & 13.5 & 24.4 & 24.2 & 4.3 & 48.8 & 45.8 & 9.8 \\
            \midrule
                     \multicolumn{13}{c}{Schedule: 3, 1, 0, 5, 4, 2}\\
            \midrule
            CRS(full)& \textcolor{blue}{26.3} & \textcolor{red}{18.6} & \textcolor{red}{21.3} & 13.5 & 14.8 & \textcolor{blue}{18.6} & 11.7 & 12.6 & 20.6 & 16.6 & 15.5 & 20.0 \\
            \;forget & 36.6 & 32.1 & 15.9 & 62.1 & 60.0 & 17.4 & 97.1 & 97.0 & 19.1 & 61.1 & 56.1 & 17.7 \\

            GSS(full)& 24.6 & 13.5 & 19.0 & \textcolor{blue}{14.8} & \textcolor{blue}{15.5} & 17.9 & 15.9 & 17.6 & 24.5 & 17.9 & 15.3 & 20.9 \\
            \;forget & 46.8 & 43.9 & 16.6 & 59.6 & 58.3 & 11.7 & 82.8 & 82.0 & 18.6 & 54.8 & 49.8 & 13.0 \\

            ExStream(full)& 15.6 & 9.2 & 15.3 & 12.4 & 12.8 & 17.6 & \textcolor{blue}{24.6} & \textcolor{blue}{24.1} & \textcolor{blue}{26.7} & \textcolor{blue}{18.7} & \textcolor{blue}{16.0} & \textcolor{blue}{21.0} \\
            \;forget & 80.7 & 77.6 & 24.0 & 81.0 & 80.6 & 23.3 & 77.2 & 76.7 & 21.8 & 81.0 & 79.3 & 23.4 \\

            PRS(full)& \textcolor{red}{26.7} & \textcolor{blue}{17.9} & \textcolor{blue}{21.2} & \textcolor{red}{19.2} & \textcolor{red}{19.3} & \textcolor{red}{21.5} & \textcolor{red}{27.5} & \textcolor{red}{26.8} & \textcolor{red}{31.0} & \textcolor{red}{24.8} & \textcolor{red}{21.7} & \textcolor{red}{25.5} \\
            \;forget & 45.8 & 43.0 & 15.7 & 59.0 & 58.4 & 13.4 & 60.6 & 60.3 & 15.5 & 55.3 & 53.5 & 13.9 \\

            \midrule
                     \multicolumn{13}{c}{Schedule: 5, 2, 0, 3, 1, 4}\\
            \midrule
            CRS(full)& \textcolor{red}{27.7} & \textcolor{red}{24.5} & \textcolor{blue}{22.5} & \textcolor{blue}{18.4} & \textcolor{blue}{14.9} & \textcolor{blue}{18.9} & 5.6 & 5.8 & 18.0 & \textcolor{blue}{16.6} & \textcolor{blue}{15.2} & 19.2 \\
            \;forget & 33.8 & 31.0 & 16.4 & 59.4 & 59.0 & 21.7 & 85.4 & 85.4 & 32.6 & 58.4 & 56.7 & 23.7 \\

            GSS(full)& \textcolor{blue}{25.1} & \textcolor{blue}{23.6} & 22.2 & 14.7 & 12.3 & 16.3 & 4.3 & 4.6 & 13.3 & 12.7 & 12.4 & 16.2 \\
            \;forget & 32.7 & 28.5 & 9.0 & 75.9 & 74.9 & 12.4 & 69.3 & 69.5 & 29.6 & 64.4 & 60.9 & 17.2 \\

            ExStream(full)& 8.4 & 9.1 & 16.2 & 17.0 & 11.2 & 17.1 & \textcolor{blue}{12.4} & \textcolor{blue}{9.4} & \textcolor{blue}{22.8} & 13.8 & 10.3 & \textcolor{blue}{19.4} \\
            \;forget & 85.8 & 84.1 & 20.6 & 78.7 & 78.2 & 24.3 & 72.1 & 71.6 & 28.7 & 80.9 & 79.5 & 25.5 \\

            PRS(full)& 23.6 & 23.3 & \textcolor{red}{23.7} & \textcolor{red}{21.6} & \textcolor{red}{16.9} & \textcolor{red}{21.8} & \textcolor{red}{23.4} & \textcolor{red}{21.9} & \textcolor{red}{30.5} & \textcolor{red}{23.2} & \textcolor{red}{20.0} & \textcolor{red}{25.9} \\
            \;forget & 57.9 & 54.5 & 10.2 & 58.9 & 58.0 & 14.5 & 42.4 & 42.6 & 13.2 & 51.8 & 49.6 & 11.8 \\
            \bottomrule
        \end{tabular}
    \end{sc}
\end{table*}

\begin{table*}
    \centering
    \scriptsize
    \setlength{\tabcolsep}{2pt}
    \caption{\textbf{\textit{COCOseq} performance on different reservoir sizes.} Please refer to table~\ref{tab:coco_ac} for the nomenclatures and details of the experiments.}
    \label{tab:coco_reservoir_size_results}
    \begin{sc}
        \begin{tabular}{lccc|ccc|ccc||ccc}
            \toprule
            &\multicolumn{3}{c|}{majority} & \multicolumn{3}{c|}{moderate}& \multicolumn{3}{c||}{minority}& \multicolumn{3}{c}{overall}\\
                     & C-F1 & O-F1 & MAP  & C-F1 & O-F1 & MAP  & C-F1 & O-F1 &  MAP & C-F1 &  O-F1 &  MAP \\
            \midrule
                     \multicolumn{13}{c}{reservoir: $1000$}\\
            \specialrule{0.3pt}{1pt}{1pt}
            CRS(full)& \textcolor{red}{63.6} & \textcolor{red}{56.2} & \textcolor{red}{64.1} & \textcolor{blue}{44.3} & \textcolor{blue}{40.2} & \textcolor{blue}{47.8} & 9.4  & 10.1 & 23.5 & \textcolor{blue}{44.2} & \textcolor{blue}{41.6}  & \textcolor{blue}{47.1} \\
            GSS(full)& 60.0 & \textcolor{blue}{55.8} & 60.2 & 40.8 & 36.9 & 42.5 & 8.2  & 8.7  & 17.6 & 40.6 & 39.0  & 42.1 \\
            ExStream(full)& 54.8 & 48.1 & 60.1 & 41.4 & 36.7 & 46.6 & \textcolor{blue}{17.7} & \textcolor{blue}{18.1} & \textcolor{blue}{31.1} & 41.6 & 37.6 & 47.0 \\
            \specialrule{0.4pt}{1pt}{1pt}
            PRS(full)& \textcolor{blue}{59.9} & 51.8 & \textcolor{blue}{62.7} & \textcolor{red}{44.6} & \textcolor{red}{40.7} & \textcolor{red}{49.7} & \textcolor{red}{26.8} & \textcolor{red}{27.5} & \textcolor{red}{34.2} & \textcolor{red}{45.9} & \textcolor{red}{42.2} & \textcolor{red}{50.0} \\
            \specialrule{0.3pt}{1pt}{1pt}
                     \multicolumn{13}{c}{reservoir: $2000$}\\
            \specialrule{0.3pt}{1pt}{1pt}
            CRS(full)& \textcolor{red}{67.0} & \textcolor{red}{62.5} & \textcolor{red}{67.9} & \textcolor{blue}{47.8} & \textcolor{blue}{45.2} & 50.4 & 14.5 & 15.6 & 26.9 & 47.5 & \textcolor{blue}{46.6}  & 50.2 \\
            GSS(full)& 59.9 & 56.7 & 60.3 & 45.4 & 43.6 & 46.4 & 10.3 & 10.7 & 18.9 & 43.2 & 43.0  & 44.4 \\
            ExStream(full)& 58.8 & 52.0 & 62.5 & 47.5 & 43.8 & \textcolor{blue}{51.0} & \textcolor{blue}{26.4} & \textcolor{blue}{26.6} & \textcolor{blue}{36.6} & \textcolor{blue}{47.8} & 43.9 & \textcolor{blue}{51.1} \\

            \specialrule{0.4pt}{1pt}{1pt}
            PRS(full)& \textcolor{blue}{65.1} & \textcolor{blue}{59.3} & \textcolor{blue}{67.2} & \textcolor{red}{51.7} & \textcolor{red}{49.5} & \textcolor{red}{55.3} & \textcolor{red}{35.0} & \textcolor{red}{34.6} & \textcolor{red}{38.3} & \textcolor{red}{52.7} & \textcolor{red}{50.1} & \textcolor{red}{55.0} \\

            \specialrule{0.3pt}{1pt}{1pt}
                     \multicolumn{13}{c}{reservoir: $3000$}\\
            \specialrule{0.3pt}{1pt}{1pt}
            CRS(full)& \textcolor{blue}{66.6} & \textcolor{red}{62.3} & \textcolor{red}{68.2} & \textcolor{blue}{49.8} & \textcolor{blue}{48.0} & 51.8 & 18.7 & 19.8 & 27.0 & 49.4 & \textcolor{blue}{48.6}  & 51.1 \\
            GSS(full)& 61.5 & 58.5 & 61.4 & 43.5 & 42.5 & 45.6 & 9.5  & 10.8 & 19.0 & 42.2 & 43.0  & 44.3 \\
            ExStream(full)& 61.7 & 53.9 & 62.7 & 49.2 & 47.3 & \textcolor{blue}{52.7} & \textcolor{blue}{30.4} & \textcolor{blue}{30.5} & \textcolor{blue}{37.6} & \textcolor{blue}{49.7} & 46.8 & \textcolor{blue}{52.2} \\
            \specialrule{0.4pt}{1pt}{1pt}
            PRS(full)& \textcolor{red}{67.1} & \textcolor{blue}{62.0} & \textcolor{blue}{68.0} & \textcolor{red}{54.1} & \textcolor{red}{52.3} & \textcolor{red}{56.4} & \textcolor{red}{36.6} & \textcolor{red}{36.8} & \textcolor{red}{39.9} & \textcolor{red}{54.6} & \textcolor{red}{52.6} & \textcolor{red}{56.1} \\
            \bottomrule
        \end{tabular}
    \end{sc}
\end{table*}

\begin{table*}
    \centering
    \scriptsize
    \setlength{\tabcolsep}{2pt}
    \caption{\textbf{\textit{COCOseq} performance on permuted schedules.} Please refer to table~\ref{tab:coco_ac} for the nomenclatures and details of the experiments.}
    \label{tab:coco_schedule_permute_results}
    \begin{sc}
        \begin{tabular}{lccc|ccc|ccc||ccc}
            \toprule
            &\multicolumn{3}{c|}{majority} & \multicolumn{3}{c|}{moderate}& \multicolumn{3}{c||}{minority}& \multicolumn{3}{c}{overall}\\
                     & C-F1 & O-F1 & MAP  & C-F1 & O-F1 & MAP  & C-F1 & O-F1 &  MAP & C-F1 &  O-F1 &  MAP \\
            \midrule
                     \multicolumn{13}{c}{Schedule: 0, 1, 3, 2}\\
            \specialrule{0.3pt}{1pt}{1pt}
            CRS(full)& \textcolor{red}{66.1} & \textcolor{red}{59.8} & 66.9 & \textcolor{blue}{50.4} & \textcolor{blue}{47.0} & \textcolor{blue}{52.2} & 14.8 & 15.4 & 26.4 & \textcolor{blue}{49.2} & \textcolor{blue}{46.9} & \textcolor{blue}{50.8} \\
            GSS(full)& 60.8 & 55.8 & 61.4 & 42.3 & 41.2 & 45.1 & 8.5 & 8.6 & 18.4 & 42.1 & 41.4 & 44.0 \\
            ExStream(full)& 60.1 & 53.0 & 62.2 & 46.6 & 42.0 & 50.6 & \textcolor{blue}{25.1} & \textcolor{blue}{25.4} & \textcolor{blue}{35.4} & 47.3 & 42.9 & 50.5 \\
            \specialrule{0.3pt}{1pt}{1pt}
            PRS(full)& \textcolor{blue}{64.4} & \textcolor{blue}{58.6} & \textcolor{red}{67.0} & \textcolor{red}{51.0} & \textcolor{red}{48.3} & \textcolor{red}{55.1} & \textcolor{red}{35.0} & \textcolor{red}{35.1} & \textcolor{red}{38.7} & \textcolor{red}{52.0} & \textcolor{red}{49.2} & \textcolor{red}{54.9} \\
            \specialrule{0.3pt}{1pt}{1pt}
                     \multicolumn{13}{c}{Schedule: 1, 0, 3, 2}\\
            \specialrule{0.3pt}{1pt}{1pt}
            CRS(full)& \textcolor{red}{66.9} & \textcolor{red}{61.1} & \textcolor{red}{67.4} & \textcolor{blue}{48.6} & \textcolor{blue}{46.7} & \textcolor{blue}{50.8} & 9.6 & 10.0 & 25.0 & \textcolor{blue}{47.9} & \textcolor{blue}{46.7} & \textcolor{blue}{49.9} \\

            GSS(full)& 59.8 & 55.7 & 60.7 & 42.6 & 41.8 & 45.2 & 11.2 & 12.2 & 19.2 & 41.5 & 41.6  & 44.0 \\

            ExStream(full)& 59.2 & 51.8 & 61.0 & 46.5 & 42.6 & 50.4 & \textcolor{blue}{24.8} & \textcolor{blue}{25.1} & \textcolor{blue}{36.0} & 46.6 & 42.9 & 50.2 \\

            \specialrule{0.4pt}{1pt}{1pt}
            PRS(full)& \textcolor{blue}{63.8} & \textcolor{blue}{57.5} & \textcolor{blue}{65.3} & \textcolor{red}{50.6} & \textcolor{red}{48.4} & \textcolor{red}{54.5} & \textcolor{red}{33.5} & \textcolor{red}{33.8} & \textcolor{red}{37.8} & \textcolor{red}{51.3} & \textcolor{red}{48.8} & \textcolor{red}{54.0} \\

            \specialrule{0.3pt}{1pt}{1pt}
                     \multicolumn{13}{c}{Schedule: 2, 3, 0, 1}\\
            \specialrule{0.3pt}{1pt}{1pt}
            CRS(full)& \textcolor{red}{65.6} & \textcolor{red}{60.1} & \textcolor{red}{66.2} & \textcolor{blue}{47.7} & \textcolor{blue}{44.0} & \textcolor{blue}{49.5} & 9.5  & 10.2 & 21.5 & \textcolor{blue}{45.4} & \textcolor{blue}{44.3} & \textcolor{blue}{48.2} \\

            GSS(full)& 54.5 & 51.1 & 58.0 & 32.6 & 32.1 & 37.6 & 4.2  & 4.1  & 15.6 & 33.5 & 33.9  & 38.5 \\

            ExStream(full)& 46.0 & 39.5 & 55.8 & 44.8 & 36.8 & 46.9 & \textcolor{blue}{25.6} & \textcolor{blue}{23.0} & \textcolor{blue}{29.5} & 41.6 & 35.2 & 45.7 \\

            \specialrule{0.4pt}{1pt}{1pt}
            PRS(full)& \textcolor{blue}{61.4} & \textcolor{blue}{56.0} & \textcolor{blue}{65.9} & \textcolor{red}{51.4} & \textcolor{red}{46.8} & \textcolor{red}{53.8} & \textcolor{red}{30.8} & \textcolor{red}{30.1} & \textcolor{red}{33.0} & \textcolor{red}{50.1} & \textcolor{red}{46.5} & \textcolor{red}{52.8} \\
            \specialrule{0.3pt}{1pt}{1pt}
                     \multicolumn{13}{c}{Schedule: 3, 1, 0, 2}\\
            \specialrule{0.3pt}{1pt}{1pt}
            CRS(full)& \textcolor{red}{65.9} & \textcolor{red}{61.0} & \textcolor{red}{66.5} & \textcolor{blue}{48.7} & \textcolor{blue}{45.1} & \textcolor{blue}{50.9} & 11.7 & 12.2 & 24.8 & \textcolor{blue}{47.7} & \textcolor{blue}{45.9} & \textcolor{blue}{49.7} \\

            GSS(full)& 59.9 & 55.8 & 58.5 & 35.8 & 35.6 & 39.3 & 11.8 & 13.5 & 18.9 & 37.8 & 38.8 & 40.2 \\

            ExStream(full)& 58.6 & 50.5 & 61.4 & 44.4 & 41.6 & 48.8 & \textcolor{blue}{24.5} & \textcolor{blue}{24.4} & \textcolor{blue}{34.7} & 45.4 & 41.8 & 49.2 \\

            \specialrule{0.4pt}{1pt}{1pt}
            PRS(full)& \textcolor{blue}{63.3} & \textcolor{blue}{57.2} & \textcolor{blue}{65.7} & \textcolor{red}{50.2} & \textcolor{red}{47.7} & \textcolor{red}{54.1} & \textcolor{red}{33.3} & \textcolor{red}{33.3} & \textcolor{red}{37.9} & \textcolor{red}{51.0} & \textcolor{red}{48.2} & \textcolor{red}{53.8} \\
            \bottomrule
        \end{tabular}
    \end{sc}
\end{table*}

\begin{table}
    \parbox{.45\textwidth}{
    \centering
    \scriptsize
    \caption{\textbf{Per task \textit{COCOseq} results}. The T=[0,1,2,3] values represent the \textit{overall} performance after the completion of each task. Please refer to table~\ref{tab:coco_ac} for the nomenclatures and details of the experiments.}
    \label{tab:coco_results}
    \begin{tabular}{lccc}
            \toprule
            & \multicolumn{3}{c}{overall}\\
                     & C-F1 &  O-F1 &  MAP \\
            \midrule
            MT       & 51.2 & 52.1  & 53.9 \\

            \midrule
            FT(full) & 8.5  & 18.4  & 16.4 \\
            \;T=0    & 48.2 & 52.8 & 55.9 \\
            \;T=1    & 19.9 & 29.2 & 34.0 \\
            \;T=2    & 12.5 & 22.1 & 22.9 \\
            \;T=3    & 8.5 & 18.4 & 16.4 \\

            forget   & 100.0& 100.0 & 70.1 \\

            \midrule
            EWC(full)& 38.9 & 40.0  & 46.6 \\
            \;T=0    & 34.1 & 39.6 & 50.5 \\
            \;T=1    & 40.4 & 45.0 & 49.6 \\
            \;T=2    & 35.8 & 38.1 & 44.4 \\
            \;T=3    & 38.9 & 40.0 & 46.6 \\
            forget   & 32.8 & 32.0  & 3.2 \\

            \midrule
            CRS(full)& 47.5 & 46.6  & 50.2 \\
            \;T=0    & 58.7 & 59.0 & 63.4 \\
            \;T=1    & 55.8 & 55.1 & 58.6 \\
            \;T=2    & 47.1 & 46.6 & 50.0 \\
            \;T=3    & 47.5 & 46.6 & 50.2 \\
            forget   & 32.2 & 30.1  & 15.3 \\

            \midrule
            GSS(full)& 42.8 & 42.7 & 44.0 \\
            \;T=0    & 55.4 & 56.0 & 59.5 \\
            \;T=1    & 53.8 & 52.8 & 55.3 \\
            \;T=2    & 42.1 & 42.5 & 44.6 \\
            \;T=3    & 42.8 & 42.7 & 44.0 \\
            forget   & 35.1 & 35.3 & 13.6 \\

            \midrule
            ExStream(full)& \textcolor{blue}{49.7} & \textcolor{blue}{46.8} & \textcolor{blue}{52.2} \\
            \;T=0    & 64.1 & 63.4 & 66.9 \\
            \;T=1    & 56.2 & 53.8 & 58.9 \\
            \;T=2    & 48.8 & 44.9 & 51.7 \\
            \;T=3    & 49.7 & 46.8 & 52.2 \\
            \;forget & 34.5 & 33.6 & 15.1 \\
            \midrule
            PRS(ours)& \textcolor{red}{53.2} & \textcolor{red}{50.3} & \textcolor{red}{55.3} \\
            \;T=0    & 61.9 & 61.8 & 65.4 \\
            \;T=1    & 59.2 & 58.1 & 61.9 \\
            \;T=2    & 53.1 & 51.1 & 55.1 \\
            \;T=3    & 53.2 & 50.3 & 55.3 \\
            \;forget & 25.6 & 25.2 & 10.2 \\
            \bottomrule
        \end{tabular}
  }
  \hfill
\parbox{.45\textwidth}{
    \centering
    \scriptsize
    \caption{\textbf{Per task \textit{NUS-WIDEseq} results}. The T=[0,1,2,3,4,5] values represent the \textit{overall} performance after the completion of each task. Please refer to table~\ref{tab:coco_ac} for the nomenclatures and details of the experiments}
    \label{tab:nus-wide_results}
    \begin{tabular}{lccc}
            \toprule
            & \multicolumn{3}{c}{overall}\\
                     & C-F1 &  O-F1 &  MAP \\
            \midrule
            MT       & 24.6 & 24.9  & 28.4 \\

            \midrule
            FT(full) & 4.2 & 5.1 & 7.1 \\

            \;T=0    & 52.6 & 50.8 & 56.6 \\
            \;T=1    & 11.2 & 16.4 & 19.3 \\
            \;T=2    & 4.9 & 9.6 & 11.1 \\
            \;T=3    & 5.5 & 9.6 & 9.9 \\
            \;T=4    & 2.2 & 6.9 & 8.4 \\
            \;T=5    & 4.2 & 5.1 & 7.1 \\

            \;forget & 100.0 & 100.0 & 44.4 \\

            \midrule
            EWC(full)& 17.1 & 11.4 & 20.7 \\
            \;T=0    & 1.3 & 1.6 & 27.0 \\
            \;T=1    & 5.8 & 9.6 & 18.9 \\
            \;T=2    & 6.7 & 8.9 & 16.3 \\
            \;T=3    & 12.5 & 12.1 & 17.5 \\
            \;T=4    & 13.5 & 13.3 & 17.9 \\
            \;T=5    & 17.1 & 11.4 & 20.7 \\

            \;forget & 36.4 & 31.5 & 7.3 \\

            \midrule
            CRS(full)& 17.5 & 15.7 & 20.3 \\
            \;T=0    & 55.8 & 53.8 & 59.1 \\
            \;T=1    & 35.8 & 27.8 & 38.0 \\
            \;T=2    & 25.0 & 21.1 & 29.1 \\
            \;T=3    & 24.9 & 22.4 & 25.8 \\
            \;T=4    & 21.6 & 20.0 & 24.5 \\
            \;T=5    & 17.5 & 15.7 & 20.3 \\
            \;forget & 60.3 & 57.3 & 17.1 \\

            \midrule
            GSS(full)& 17.9 & 15.3 & 20.9 \\
            \;T=0    & 55.2 & 53.2 & 56.9 \\
            \;T=1    & 43.9 & 39.6 & 43.3 \\
            \;T=2    & 33.8 & 29.3 & 34.2 \\
            \;T=3    & 28.1 & 26.4 & 29.1 \\
            \;T=4    & 22.3 & 21.5 & 27.0 \\
            \;T=5    & 17.9 & 15.3 & 20.9 \\
            \;forget & 54.8 & 49.8 & 13.0 \\

            \midrule
            ExStream(full)& \textcolor{blue}{18.7} & \textcolor{blue}{16.0} & \textcolor{blue}{21.0} \\
            \;T=0    & 57.6 & 55.8 & 59.9 \\
            \;T=1    & 33.9 & 24.4 & 38.3 \\
            \;T=2    & 23.0 & 17.6 & 28.7 \\
            \;T=3    & 21.8 & 17.3 & 26.3 \\
            \;T=4    & 18.2 & 14.6 & 22.6 \\
            \;T=5    & 18.7 & 16.0 & 21.0 \\
            \;forget & 81.0 & 79.3 & 23.4 \\

            \midrule
            PRS(full)& \textcolor{red}{24.8} & \textcolor{red}{21.7} & \textcolor{red}{25.5} \\
            \;T=0    & 56.1 & 54.3 & 58.7 \\
            \;T=1    & 44.3 & 38.5 & 44.3 \\
            \;T=2    & 34.3 & 29.0 & 35.2 \\
            \;T=3    & 32.5 & 30.2 & 32.5 \\
            \;T=4    & 30.8 & 28.0 & 30.5 \\
            \;T=5    & 24.8 & 21.7 & 25.5 \\
            \;forget & 55.3 & 53.5 & 13.9 \\
            \bottomrule
        \end{tabular}
}
\end{table}

\end{document}